\documentclass[iicol]{sn-jnl}

\usepackage{graphicx}%
\usepackage{multirow}%
\usepackage{amsmath,amssymb,amsfonts}%
\usepackage{amsthm}%
\usepackage{mathrsfs}%
\usepackage[title]{appendix}%
\usepackage{textcomp}%
\usepackage{manyfoot}%
\usepackage{booktabs}%
\usepackage{algorithm}%
\usepackage{listings}%
\usepackage{hyperref}
\usepackage{url}
\usepackage{amsmath}
\usepackage{algorithm}
\usepackage{algorithmic}
\usepackage{tabularx,ragged2e}
\usepackage{pifont}
\newcommand{\cmark}{\ding{51}}%
\newcommand{\xmark}{\ding{55}}%
\usepackage{booktabs}
\usepackage{booktabs}
\usepackage{adjustbox,makecell}
\usepackage{amsmath,amsfonts}
\usepackage{algorithmic}
\usepackage{algorithm}
\usepackage{array}
\usepackage{textcomp}
\usepackage{stfloats}
\usepackage{url}
\usepackage{verbatim}
\usepackage{graphicx}
\usepackage{amsmath}
\usepackage{amssymb}
\usepackage{booktabs}
\usepackage{adjustbox}
\usepackage{booktabs}
\usepackage{breakurl}

\usepackage{siunitx}
\usepackage{amsmath}
\usepackage{algorithm}
\usepackage{algorithmic}
\usepackage[bb=px]{mathalfa} 
\usepackage{mathabx}
\usepackage{epstopdf}
\usepackage{multirow}
\usepackage{siunitx}
\usepackage{soul}
\usepackage[accsupp]{axessibility}
\usepackage[absolute,overlay]{textpos}
\usepackage{subcaption}
\usepackage{amsmath}
\usepackage[table,xcdraw]{xcolor}
\usepackage[font=small]{caption}
\usepackage{algorithm}
\usepackage{amssymb} 
\usepackage{tikz}
\usepackage{subcaption}
\usepackage{amsmath}
\usetikzlibrary{arrows.meta,positioning,fit,calc,shapes.misc,shadows.blur}
\usepackage{tabularx}
\usepackage{graphicx}  
\usepackage{colortbl}  
\usepackage{booktabs}  
\usepackage{xspace}
\usepackage{array}
\usepackage[inline]{enumitem}
\usepackage{caption}  
\usepackage{colortbl} 
\usepackage{booktabs}
\usepackage{xcolor}
\usepackage{float}

\newenvironment{blueblock}{
  \color{blue} 
}{}

\theoremstyle{thmstyleone}%
\theoremstyle{thmstyletwo}%

\theoremstyle{thmstylethree}%

\begin{document}

\title[ ]{Context Aware Grounded Teacher for Source Free Object Detection}


\author*[1,3]{\fnm{Tajamul} \sur{Ashraf}}\email{tajamul.ashraf@mbzuai.ac.ae}

\author[3]{\fnm{Rajes} \sur{Manna}}
\equalcont{These authors contributed equally to this work.}

\author[2]{\fnm{Partha Sarathi} \sur{Purkayastha}}
\equalcont{These authors contributed equally to this work.}

\author[3]{\fnm{Tavaheed } \sur{Tariq}}

\author*[3]{\fnm{Janibul} \sur{Bashir}}\email{janibbashir@nitsri.ac.in}

\date{Accepted in International Journal of Computer Vision (IJCV)}

\affil*[1]{\orgdiv{Department of Computer Vision}, \orgname{MBZUAI}, \orgaddress{\city{Masdar City}, \postcode{11058}, \state{Abu Dhabi}, \country{UAE}}}
\affil*[2]{\orgname{Microsoft Research India}, \orgaddress{\city{Bengaluru}, \postcode{560001}, \state{Karnataka}, \country{India}}}

\affil*[3]{\orgdiv{GAASH Research Lab}, \orgname{NIT Srinagar}, \orgaddress{\street{Hazratbal}, \postcode{190007}, \state{J\&K}, \country{India}}}


\abstract{

Source-free object detection (SFOD) faces persistent challenges due to class imbalance–driven context bias and instability in teacher–student training under noisy pseudo-labels. Existing techniques tend to ignore the context bias and class imbalance shift, especially in medical data. To tackle this, we propose Grounded Teacher (\gt), a bias-aware source-free framework that grounds the teacher model through relational and semantic regularization.. To explicitly model directional confusions between classes, \gt introduces a Relational Context Module (RCM), which maintains an exponential moving average (EMA) estimate of cross-domain contextual bias. Building upon this, a Semantic Augmentation (SA) strategy selectively augments minority and confusable classes through adaptive MixUp in both source-similar and source-dissimilar target regions, improving minority recall without overfitting dominant categories. To stabilize learning under biased pseudo-labels, we design a Semantic-Aware Loss (SAL) that applies diagonally normalized weights, preventing gradient explosion while emphasizing minority majority corrections. Additionally, a frozen Expert branch derived from large vision foundation models (LVFMs) serves as a supervisory reference during training, refining pseudo-label quality without adding inference overhead. GT’s behavior-driven bias quantification makes it broadly applicable across domains without relying on dataset priors. Evaluations on \textbf{Cityscapes→Foggy} (50.8 mAP) and medical transfers (\textbf{+5.9 AP\textsubscript{50}} on DDSM→INBreast) show consistent gains and improved minority-class detection, with less than 12\% additional training cost. Code and model are available at \href{https://github.com/Tajamul21/Grounded_Teacher}{https://github.com/Tajamul21/Grounded-Teacher}.
}

\keywords{Generalization, Source-Free Domain Adaptation, Context Bias, Semi-Supervised Learning, Large Vision Foundation Models, Knowledge Distillation}


\def\sfod{\texttt{SFOD}\xspace}
\def\gt{\texttt{GT}\xspace}
\def\uda{\texttt{UDA}\xspace}
\def\st{\texttt{ST}\xspace}
\def\ema{\texttt{EMA}\xspace}
\def\rcm{\texttt{RCM}\xspace}
\def\vfms{\texttt{VFMs}\xspace}
\def\sota{\texttt{SOTA}\xspace}
\def\edf{\texttt{EDF}\xspace}
\def\frcnn{\texttt{FRCNN}\xspace}
\def\lods{\texttt{LODS}\xspace}
\def\aasfod{\texttt{AASFOD}\xspace}
\def\irg{\texttt{IRG}\xspace}
\def\soap{\texttt{SOAP}\xspace}
\def\pets{\texttt{PETS}\xspace}
\def\afal{\texttt{AFAL}\xspace}
\def\kd{\texttt{KD}\xspace}
\def\fm{\texttt{FM}\xspace}
\def\clip{\texttt{CLIP}\space}
\def\dinov{\texttt{DINOV}\xspace}
\def\sam{\texttt{SAM}\xspace}
\def\ckd{\texttt{CKD}\xspace}
\def\icrm{\texttt{ICRm}\xspace}
\def\fifo{\texttt{FIFO}\xspace}
\def\vit{\texttt{ViT}\xspace}
\def\lvfms{\texttt{LVFMs}\xspace}
\def\laion400m{\texttt{LAION-400M}\xspace}
\def\sfda{\texttt{SFDA}\xspace}
\def\bcd{\texttt{BCD}\xspace}
\def\mi{\texttt{MI}\xspace}
\def\cf{\texttt{CF}\xspace}
\def\cb{\texttt{CB}\xspace}
\def\kc{\texttt{KC}\xspace}
\def\sc{\texttt{SC}\xspace}
\def\ii{\texttt{II}\xspace}
\def\id{\texttt{ID}\xspace}
\def\fnd{\texttt{FND}\xspace}
\def\ap{\texttt{AP}\xspace}
\def\map{\texttt{mAP}\xspace}
\def\froc{\texttt{FROC}\xspace}
\def\fpi{\texttt{FPI}\xspace}
\def\auc{\texttt{AUC}\xspace}
\def\kitti{\texttt{KITTI}\xspace}
\def\nvidia{\texttt{NVIDIA}\xspace}
\def\a100{\texttt{A100}\xspace}
\def\gpus{\texttt{GPUs}\xspace}
\def\ce{\texttt{CE}\xspace}
\def\coco{\texttt{COCO}\xspace}
\def\pascal{\texttt{PASCAL}\xspace}
\def\wsod{\texttt{WSOD}\xspace}
\def\vit{\texttt{ViT}\xspace}
\def\detr{\texttt{DETR}\xspace}
\def\daod{\texttt{DAOD}\xspace}
\def\di{\texttt{DI}\xspace}
\def\sfdaod{\texttt{SFDOAD}\xspace}
\def\HT{\texttt{HT}\xspace} 
\def\umt{\texttt{UMT}\xspace}
\def\tdo{\texttt{TDO}\xspace}
\def\pt{\texttt{PT}\xspace}
\def\at{\texttt{AT}\xspace}
\def\cmt{\texttt{CMT}\xspace}
\def\mila{\texttt{MILA}\xspace}
\def\mrt{\texttt{MRT}\xspace}
\def\cra{\texttt{CRA}\xspace}
\def\icl{\texttt{ICL}\xspace}
\def\iou{\texttt{IoU}\xspace}
\def\sal{\texttt{SAL}\xspace}
\def\sa{\texttt{SA}\xspace}
\def\sl{\texttt{SL}\xspace}
\def\cat{\texttt{CAT}\xspace}
\def\mil{\texttt{MIL}\xspace}
\def\fcos{\texttt{FCOS}\xspace}
\def\dbt{\texttt{DBT}\xspace}
\def\yolov5{\texttt{YOLOv5}\xspace}
\def\btmuda{\texttt{BTMuda}\xspace}
\def\ct{\texttt{CT}\xspace}
\def\sfa{\texttt{SFA}\xspace}
\def\h2fa{\texttt{H2FA}\xspace}
\def\aqt{\texttt{AQT}\xspace}
\def\dadapt{\texttt{D-Adapt}\xspace}
\def\clipgap{\texttt{CLIPGAP}\xspace}
\def\confmix{\texttt{ConfMIX}\xspace}
\def\dmaster{\texttt{D-MASTER}\xspace}
\def\mexformer{\texttt{Mexformer}\xspace}
\def\mammnet{\texttt{MAMM-Net}\xspace}
\def\mask2former{\texttt{Mask2Former}\xspace}
\def\png{\texttt{PNG}\xspace}
\def\lpld{\texttt{LPLD}\xspace}
\def\jpeg{\texttt{JPEG}\xspace}
\def\groundingdino{\texttt{GroundingDINO}\xspace}
\def\biomedparse{\texttt{BioMedParse}\xspace}
\def\daca{\texttt{DACA}\xspace}

\maketitle

\small Accepted in International Journal of Computer Vision

\clearpage

\section{Introduction}
\label{sec:intro}

Medical object detection is a well-studied problem in computer vision~\cite{carion2020end, dosovitskiy2020image, zhu2020deformable, zhang2022dino, yang2022focal}. The success of deep learning techniques for the problem has been supported by the abundance of extensively annotated detection datasets \cite{cordts2016cityscapes, everingham2010pascal, geiger2013vision, lin2014microsoft, yu2020bdd100k}, which facilitates the supervised training of deep object detection models. However most of the medical data is not available due to policy and privacy reasons.
Unsupervised Domain Adaptation (\uda) offers a strategy for adapting object detectors to new domains where labeled data is not available.

This arises from the growing reliance on medical data, where annotations are not only expensive to obtain but also prone to errors in complex or ambiguous scenarios. It has been widely observed that deep object detection models often struggle to generalize to new visual domains despite their effectiveness in familiar visual contexts. \uda is a popular solution strategy \cite{chen2017no, ganin2016domain, hoffman2018cycada, hoffman2016fcns, lo2022learning, saito2018maximum, tzeng2017adversarial}, which bridges the gap between source and target domains by aligning the feature distributions \cite{chen2018domain, he2019multi, inoue2018cross, saito2019strong, vs2023instance, huang2022aqt, zhao2023masked}. However, the strategy requires access to source-domain data at the adaptation stage, which severely limits its applicability in medical imaging \cite{liang2020we, liu2021source, xia2021adaptive}. This motivates us to focus on Source-Free Object Detection (\sfod) in this work.

Source-free domain Adaptation (\texttt{SFDA}) has received significant attention for the image classification task in recent years \cite{yang2022attracting, zhang2022divide, wang2022metateacher, jing2022variational, dong2021confident, huang2021model, yang2021exploiting}. However, there are relatively fewer works specifically addressing \sfod \cite{huang2021model,vs2023towards, oza2023unsupervised, liu2023periodically, chu2023adversarial}. Given the complexities of cluttered backgrounds, viewpoint variations, and a large number of negative samples in medical images, directly applying conventional \sfod methods for classification tasks to object detection in medical imaging leads to unsatisfactory performance. As these methods ignore the critical issue of context imbalance, which is a problem in many real-life scenarios, such as medical imaging.  Thus, there is a need to develop \sfod techniques specially tailored for medical imaging due to clinical relevance~\cite{ashraf2024dmastermaskannealedtransformer}.
\begin{figure}[t]
    \centering
    \includegraphics[width=\columnwidth]{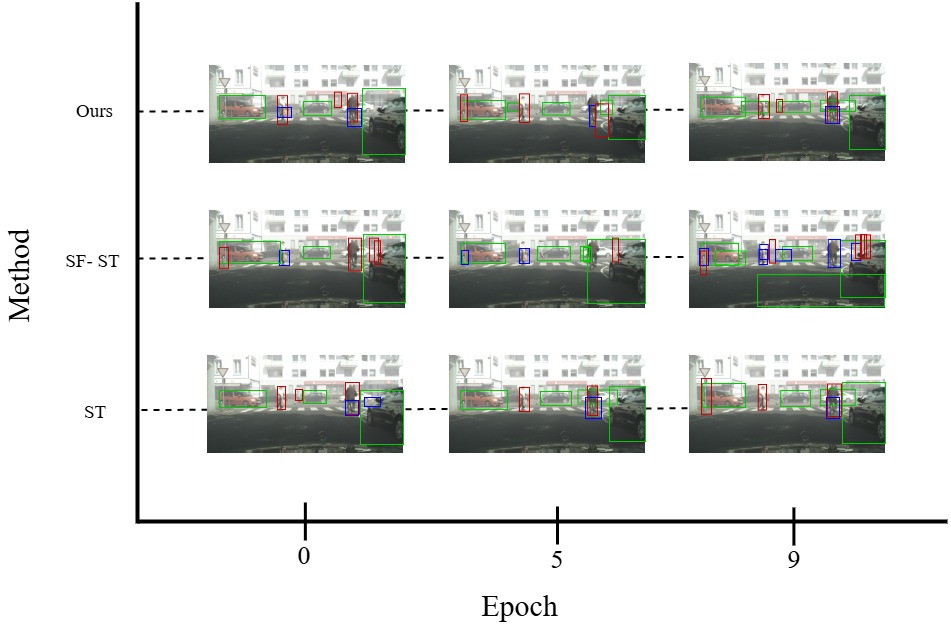}
    \caption{\textbf{Effect of pseudo-label noise and EMA drift.} 
    Visualization of target-domain detections (Cityscapes$\rightarrow$Foggy) over training epochs for (a) Mean Teacher (ST), (b) its source-free variant (SF-ST), and (c) Grounded Teacher (GT). 
    Our framework maintains spatially consistent, low-noise predictions through its RCM, SA/SAL, and expert supervision, mitigating drift and improving stability.}
    \label{fig:degradation-label}
\end{figure}

Moreover, many popular \sfod techniques utilize a self-supervised approach in a student-teacher (\st) framework. These methods bootstrap by training on pseudo-labels generated by a source-pretrained model ~\cite{li2022source, xiong2021source,li2021free, chu2023adversarial, vs2022mixture}. However, if the source data is biased or the domain shift between source and target domains is significant, there is noise in the pseudo-labels, which impacts the training of a student model \cite{deng2021unbiased}. Since the pseudo-label error is significant, the Exponential Moving Average (\ema) step, which updates the teacher model from the student model's weight, ends up corrupting the teacher model as well, leading to an exploding grading and model collapses~\cite{ali2024cftsgancontinualfewshotteacher}. This phenomenon of progressive degradation, especially in source-free settings where source data is not available during adaptation, is visually evident in Figure~\ref{fig:degradation-label}, where traditional Mean Teacher frameworks and SF-Student-Teacher frameworks suffer from declining detection quality over time, while our method remains robust over the epochs.

Prior works have documented pseudo-label noise and EMA instabilities in teacher–student adaptation, including in SFOD baselines and variants inspired by Unbiased Teacher/Mean-Teacher families~\cite{vs2023instance}. Our contribution is not simply the observation of these issues, but a coherent, source-free solution to address them. In this paper, we introduce our Grounded Teacher (\gt), specifically designed to tackle context imbalance and mode collapse in the \sfod setting, which comprises a behavioral confusion estimator (RCM) with dual use in data-space augmentation (SA) and loss-space reweighting (SAL), along with training-only LVFM Expert guidance to stabilize the teacher. Together, these mechanisms target context bias and teacher drift without any source images or inference-time cost. This is typically not a problem in the \uda setting, as supervised data from the source domain acts as an anchor and prevents error accumulation in the student. Then the \ema step ensures that the teacher model does not get corrupted at any point in the adaptation.

The Relation-Contextual Module (\rcm) explicitly models class biases and inter-class dynamics within the training data. Leveraging prior insights, we implement a semantic augmentation strategy that enhances minority class representation by strategically blending them with visually similar majority class instances, drawn from our Cropbank repository\cite{Zhang_Pan_Wang_2022}. This augmentation operates bidirectionally across domains, simultaneously addressing both class imbalance and domain shift. To further mitigate inter-class bias, we propose a Semantic-Aware Loss that dynamically prioritizes minority classes during training, with a particular focus on challenging cases where they are most vulnerable to misclassification compared to majority classes. Together, these components form a comprehensive framework that holistically addresses the challenges of representation learning in cross-domain scenarios. 

 Another key insight of this paper is that one can use learnt representations from Vision Foundational Models to guide a student model in \sfod settings. Due to the excellent zero-shot capabilities of the Large Vision Foundation Models (\lvfms), they can effectively guide student models to learn the right representation and distances between the samples, even when the target domain has a large shift from the source domain. This helps the model overcome mode collapse caused by biased pseudo-labels during unsupervised training.

Through the integration of these approaches, we observe enhanced pseudo-label quality, leading to measurable gains on medical imaging benchmarks, achieving a new state-of-the-art (\sota) of 50.8  \map in natural images for Cityscapes $\to$ Foggy Cityscapes, beating the previous standard of 45.0 \map by +5.8 \map ($\sim$13\% performance gain).
We summarize the contributions of this paper as follows:
\begin{itemize}
    \item We highlight that minority-class samples are particularly prone to pseudo-label noise and source bias in \sfod, leading to unstable student updates and degraded detection performance under strong domain shifts.
   \item We propose our \gt model, supported by our \rcm, which can map the model's existing class biases. To this end, we formulate a Semantic Aware Loss (\sal) to further enhance the performance of biased classes.
  
   \item We propose to leverage the zero-shot performance of \vfms to guide the student model in a sample-specific manner and help learn the correct representation in a large domain shift scenario.

    \item We observe that \sfod formulation is especially used for medical imaging scenarios where sharing source data is harder due to stricter privacy constraints, and where source data bias is more likely due to smaller datasets captured in a single center mode. Hence, we experiment and demonstrate \sota results on cross-domain datasets for Breast Cancer Detection. We report a recall of 0.45 and 0.43  at an \fpi of 0.3 on publicly available DDSM to INBreast~\cite{moreira2012inbreast} and DDSM~ to RSNA datasets, respectively, against the performance of 0.05 and 0.31  by the current \sota \cite{vs2023instance}.

\end{itemize}

\section{Related Work}
\label{related}

\noindent\textbf{Object Detection.}
In recent years, object detection has become a central task in computer vision, garnering significant attention due to its wide range of applications \cite{cheng2023towards, zou2023object}. Traditional object detectors typically perform bounding box regression and category classification using different paradigms such as anchors \cite{redmon2016you, lin2017focal, redmon2017yolo9000}, region proposals \cite{ren2015faster, girshick2015fast}, and point-based methods \cite{wang2017point, zhou2019objects, tian2022fully}. These approaches often rely on large-scale datasets like COCO \cite{lin2014microsoft} and PASCAL \cite{everingham2010pascal}, which require extensive manual annotation and substantial data per object category.
To reduce annotation costs, Weakly Supervised Object Detection (\wsod) methods have emerged as a promising alternative \cite{zhang2020weakly, zhang2021weakly}. These methods utilize image-level labels \cite{ma2024codet} instead of instance-level annotations, treating each image as a bag of proposals and leveraging Multiple Instance Learning (\mil) \cite{bilen2015weakly, cinbis2016weakly, bilen2016weakly, wang2023alwod} to assign labels to regions of interest. This cost-effective approach enables expansion of the detection vocabulary using only classification-level data \cite{deng2009imagenet, zhou2022detecting, wang2023alwod, kniesel2023weakly, gungor2024boosting, wang2024weakly}.
More recently, transformer-based models \cite{ashish2017attention} have gained prominence in object detection due to their ability to model token-wise dependencies and contextual relationships. Inspired by the Vision Transformer (\vit) \cite{dosovitskiy2020image} and Detection Transformer (\detr) \cite{carion2020end}, several transformer-based detectors have been proposed \cite{carion2020end, zhu2020deformable, zheng2020end, dai2021up, zhang2022dino, yang2022focal, ashraf2024hffedhierarchicalbasedcustomized}, offering state-of-the-art performance across benchmarks.
In biomedical imaging, detection techniques are increasingly adopted for critical tasks such as cancer diagnosis. Varlamova et al. \cite{varlamova2024featuresfusiondualviewmammography} proposed \mammnet, based on \mask2former, for fusing dual-view mammography features to localize malignant lesions, mimicking a radiologist’s diagnostic reasoning. Wang et al. \cite{33fea9ba3b6c428ba1d6a6d2af4817d3} utilized an anchor-free \fcos network with segmentation-guided enhancement for lesion detection in ultrasound images. Alashban \cite{Alashban2024} employed a CBAM-enhanced \yolov5 for breast cancer detection in Digital Breast Tomosynthesis (\dbt) images, while Kebede et al. \cite{Kebede2024} also used \yolov5 to identify suspicious masses in mammograms, emphasizing interpretability and diagnostic assistance. Yan et al. \cite{yan2019mulanmultitaskuniversallesion} extended Mask R-CNN to detect lesions in \ct scans using bounding boxes. Outside detection, DRSL (Distribution Regularized Self-Supervised Learning)~\cite{iqbal2022distribution} regularizes
 pixel-level multi-modal class distributions for domain-adaptive segmentation and aligns source+target
 modes to reduce pseudo-label noise. DRSL is informative because it also addresses distributional noise,
 but operates with a different objective: to minimize intra-class distribution mismatch between source and
 target domains at the pixel level by aligning multi-modal latent features for each class, while preserving
 class consistency across domains. In contrast, GT’s objective operates instance-wise and source-free,
 targeting inter-class confusion reduction rather than cross-domain distribution alignment. Our detec
tor estimates class-wise confusion from target pseudo-labels, routes augmentation through variance
split Cropbanks, and applies a diagonal-normalized, bounded loss that emphasizes minority→majority
 confusions without gradient blow-up. The two perspectives are therefore complementary rather than
 interchangeable across tasks.
Despite these advances, most approaches remain reliant on supervised learning. In this work, we aim to improve model adaptability across domains without requiring additional annotations.

\noindent\textbf{Unsupervised Domain Adaptation (\uda).}
\uda seeks to reduce the gap between domains by utilizing unlabeled target domain samples in object detection tasks. It is particularly effective when labeled data is accessible in the source domain for adaptation. Chen et al. \cite{chen2018domain} pioneered \daod, exploring adversarial feature alignment for Faster R-CNN (\frcnn) \cite{frcnn}. Subsequent works \cite{he2019multi, chen2018re, kim2019diversify,  chen2020structure, chen2021i3net, rezaeianaran2021seeking, xu2020cross} have investigated various aspects of feature alignment, while others \cite{wang2021exploring, huang2022aqt, gong2022improving,  zhao2023masked} have elaborately designed alignments for transformer architectures. Among these methodologies, approaches employing the student-teacher (\st) framework have gained prominence due to their superior performance in experimental evaluations. For example, Li et al. \cite{li2022cross} employed adversarial alignment and weak-strong augmentation techniques to reduce the false positive ratio of pseudo-labels, while He et al. \cite{he2022cross} mitigated domain shift by incorporating target object knowledge through self-distillation. Kennerley et al. \cite{kennerley2024catexploitinginterclassdynamics} introduced the Class-Aware Teacher (\texttt{CAT}) framework, employing an Inter-Class Relation module (\texttt{ICRm}) to model class biases and using Class-Relation Augmentation (\texttt{CRA}) with an Inter-Class Loss (\texttt{ICL}) to mitigate class imbalance, demonstrating improved performance on \texttt{UDA} benchmarks. Additionally, Zhao et al. \cite{zhao2023masked} utilized mask image modeling to align source and target domains and learn domain-invariant features. \uda has also been explored for medical image applications. Ryan et al. \cite{ryan2021unsupervised} used adversarial \uda with a U-Net architecture to segment breast tissue in mammograms from different acquisition devices. \uda has also been used for cell detection in digital pathology, where Pina et al. \cite{pina2024unsupervised} utilized adversarial feature alignment with detection transformers to address cross-stain generalization in breast cancer analysis. It was also actively explored for breast cancer detection from mammograms, as demonstrated by Ashraf et al. \cite{ashraf2024dmastermaskannealedtransformer} who introduced \dmaster, a transformer-based \st framework with adaptive masking and confidence refinement. \uda has also been extended to leverage multiple source domains, particularly in medical image analysis, as seen in the work by Yang et al. \cite{yang2024btmudabilevelmultisourceunsupervised} who introduced \btmuda, a bi-level multi-source \uda framework for breast cancer diagnosis using mammograms. Despite their significant progress, these methods heavily rely on access to source data, which can be challenging to obtain in real-life scenarios due to privacy concerns and data constraints. In the absence of source data, these methods encounter challenges with low-quality pseudo-labels, often resulting in a limited number of bounding boxes and false positives. To address these issues, we build upon the baseline of the \st framework with Exponential Moving Average (\ema), introducing a foundation-based Expert branch and semantic-aware loss to address these issues.

\noindent\textbf{Source-Free Object Detection (\sfod).}
In real-world applications, access to source data during the adaptation phase is often limited due to privacy concerns, data transmission limitations, or proprietary restrictions, especially in medical imaging. \sfod refers to the approach of adapting a pre-trained source model to a target domain without needing access to the source data \cite{luo2024crots}. Due to these constraints, \sfod has become more challenging compared to traditional Unsupervised Domain Adaptation  (\uda) methods, leading to the emergence of \sfod as a distinct branch within the domain adaptation framework in recent years. Given the intricate nature of object detection tasks, which involve numerous regions, multi-scale features, and complex network architectures, coupled with the absence of source data and target pseudo-labels, it is evident that the straightforward application of existing \uda methods is inadequate. The proposed \sfod in \cite{li2021free} uses self-entropy descent to create high-quality pseudo-labels for self-training. \soap \cite{xiong2021source} introduces domain perturbation on the target data, aiding the model in acquiring domain-invariant features resilient to the perturbations. \lods \cite{li2022source} introduces a style enhancement module and a graph alignment constraint to facilitate the model in learning domain-independent features. \aasfod~\cite{chu2023adversarial} categorizes target images into source-similar and source-dissimilar images, followed by employing adversarial alignment with \st models. \irg~\cite{vs2023instance} introduces an instance relation graph network integrated with contrastive loss to facilitate contrastive representation learning. A close line of work in domain-adaptive object detection (DAOD) argues that inter-class relations can be exploited to counter class bias when both source and target images are available. CAT~\cite{kennerley2024catexploitinginterclassdynamics}
 (Class-Aware Teacher) learns an Inter-Class Relation module, uses those relations to drive class-relation
 augmentation, and introduces a class-relation loss. The approach is trained with source data and re
ported on Cityscapes→Foggy and related shifts, where the results support the premise that explicitly
 modeling relations helps close minority-class gaps, but its mechanisms assume access to source super
vision during adaptation.
 By contrast, IRG~\cite{vs2023instance} treats source-free detection and builds an Instance Relation Graph on target
 proposals to guide a contrastive objective inside a mean-teacher loop. IRG’s graph, therefore, is the
 supervisory signal that shapes representations, without performing confusion-conditioned augmentation
 or class-wise bounded reweighting. Our method remains in the source-free regime like IRG, but it differs
 in what is modeled (a behavioral, directed confusion matrix rather than a latent graph), how it is used
 (to steer MixUp pairing and loss weights), and where cost is paid (LVFM guidance only at training time,
 no inference overhead). \pets \cite{liu2023periodically} introduces a dynamic teacher within the \st framework and proposes a consensus mechanism to integrate predictions from both the static teacher and the dynamic teacher models. While works in medical imaging have not explicitly used \sfod, they have utilized techniques that reflect the fundamental ideals of \sfod like Yan et al. \cite{yan2019mulanmultitaskuniversallesion} use Mask R-CNN and pseudo-masks, Akkus et al. \cite{akkus2017deep} discuss transfer learning in segmentation, Wang et al. \cite{wang2022breast} use auxiliary segmentation, various works \cite{Kebede2024, Alashban2024, khalid2023breast} employ advanced detection architectures and transfer learning for medical image analysis. These approaches highlight the principles of adapting models for detection tasks in new contexts, mirroring the goals of \sfod. Despite leveraging the \st framework, most of these approaches often overlook the challenge of overcoming local optima and training instability resulting from a single updated teacher model. To address this, we propose a Grounded Teacher framework. It leverages knowledge from expert models to mitigate catastrophic forgetting and uses a relation contextual module for generating refined pseudo-labels by removing context bias.

\section{Preliminaries}

\textbf{Problem statement. }Let $X$ denote the input space and $\mathcal{D}_s = \{(x_s^i, y_s^i) \}_{i=1}^{N_s}$, the source domain, where $y_s^i = (\mathbf{b_s^i}, c_{s}^i)$ is a tuple comprising of ground truth bounding box ($\mathbf{b_s^i} \in \mathbf{\mathcal{R}^4}$) and the corresponding class label ($c_{s}^i \in \mathbf{\mathcal{R}}$) of the object respectively. Here, $x_{s} \in X$ denotes the image from the source domain, and $N_s$ denotes the total number of source images. The target domain $\mathcal{D}_t = \{x_{t}^i\}_{i=1}^{N_t}$ is unlabeled, where $N_t$ denotes the total number of target images, and the target samples $x_{t}^i \in X$. Our objective is to transfer the source-trained model to the target domain without accessing samples from the source domain.

\noindent\textbf{Mean Teacher Framework.} 
The self-supervised adaptation strategy involves updating the student model using unlabeled target data, leveraging pseudo-labels generated from the teacher model. Both the student and the teacher models are initialized by the source-trained model. The pseudo labels undergo a confidence threshold-based filtering process, and only the reliable ones are utilized to supervise the student training \cite{khodabandeh2019robust}. The pseudo-label supervision loss for the object detection model can be expressed as:
\begin{equation}
    \mathcal{L}_\text{stu} = \mathcal{L}_\text{box}^S (x_t^i, yt_t^i) + \mathcal{L}_\text{giou}^S (x_t^i, yt_t^i) + \mathcal{L}_\text{cls}^S (x_t^i, yt_t^i).
    \label{eq1}
\end{equation}
Here $yt_t^i$ represents the pseudo-label for image $i$ obtained after filtering low-confidence predictions from the teacher. The teacher is updated progressively through an Exponential Moving Average (\ema) of student weights.  We implement a hard threshold, $\tau$, on the classification scores generated by the teacher to ensure that only pseudo-labels with high confidence are utilized by the student network, thereby encouraging more reliable learning outcomes. 
Thus, the overall self-training process for the traditional student-teacher (\st) based object detection framework is expressed as follows:
\begin{align}
\Theta_s &\leftarrow \Theta_s + \gamma \frac{\partial L_{\text{stu}}}{\partial \Theta_s}, \tag{2} \\
\Theta_t &\leftarrow \alpha \Theta_t + (1 - \alpha) \Theta_s. \tag{3}
\end{align}

Here $\mathcal{L}_\text{stu}$ represents the student loss computed using pseudo-labels from the teacher network, and $\Theta_s$ and $\Theta_t$ denote the student and teacher network, respectively. The parameters $\gamma$ and $\alpha$ denote the student learning rate and teacher \ema rate, respectively. Although the \st framework enables knowledge distillation using noisy pseudo-labels, it alone is insufficient for learning high-quality target features in a source-free setting and leads to model degradation. Therefore, to \texttt{enhance} the features in the target domain, we introduce Visual Foundation Models (\vfms) as \texttt{Expert} and incorporate supervised loss for knowledge transfer from expert \vfms. Following \cite{at}, a discriminator is added to encourage domain invariant feature representations with an associated loss, $\mathcal{L}_{\rm dis}$.

\begin{figure*}
    \centering
    \includegraphics[width=\linewidth]{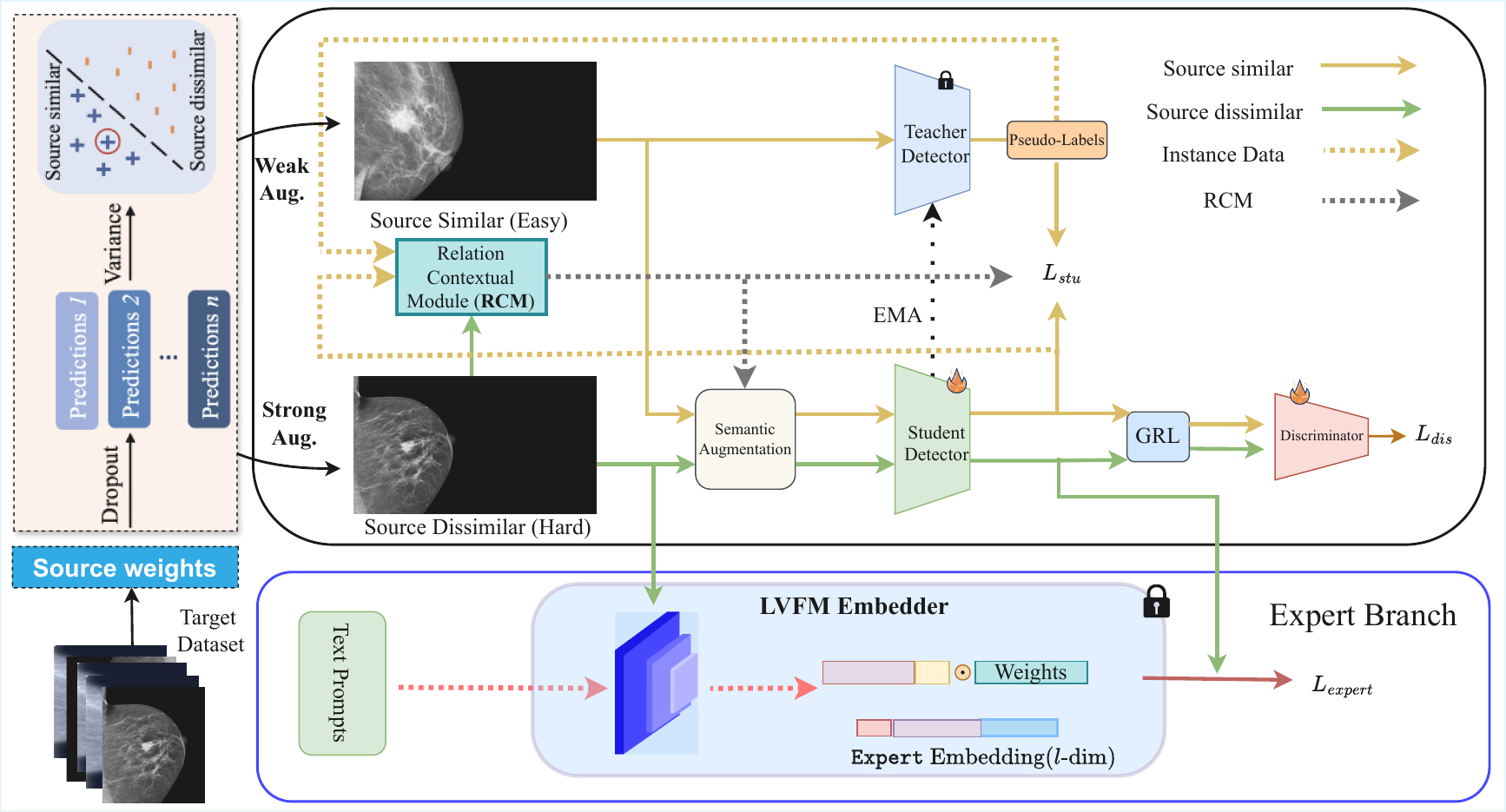}
    \caption{Architecture of proposed Grounded Teacher (\gt) Framework}
    \label{main_arch}
\end{figure*}
\section{Methodology}
\label{sec:method}

Our proposed method, Grounded Teacher (\gt), illustrated in Figure~\ref{main_arch}, extends the standard mean-teacher framework by introducing a novel Relation Contextual Module (\rcm). At the heart of \gt, the \rcm is designed to systematically capture and quantify class-specific biases in the model. Unlike conventional methods that address class imbalance in a general way, \rcm explicitly models the semantic relationships between classes, with a particular focus on minority classes that are frequently misclassified as dominant ones.

This is achieved through the construction of a batch-wise confusion matrix based on pseudo-labels, which is normalized and continuously aggregated into a global matrix. This evolving matrix provides a robust and dynamic representation of the model’s bias landscape, forming the foundation for two key components of our method: Class-Relation Augmentation and Semantic Loss. Class-Relation Augmentation targets image-level class imbalance. Using the class relationships encoded by \rcm, we identify minority classes in an image that share strong semantic overlap with dominant classes. To increase their representation and improve discrimination, we apply MixUp \cite{mixup}, a technique known to be effective in imbalanced scenarios \cite{balmixup,remix}, to blend instances of these semantically related classes. This not only amplifies the presence of minority class samples but also encourages the model to learn finer distinctions between closely related categories (Figure~\ref{main_arch}).

In parallel, the \rcm guides the formulation of our Semantic-Aware Loss, a weighted loss function that assigns greater importance to minority classes—especially those prone to misclassification. By amplifying the penalty for errors on these classes, our approach effectively mitigates the model’s inherent biases and promotes balanced, equitable learning across all categories.

\subsection{Relation Contextual module}
Previous efforts in addressing class imbalance within domain adaptation settings \cite{hsuUnsupervisedDomainAdaptation2015, jiangImplicitClassConditionedDomain2020, tanwisuthPrototypeOrientedFrameworkUnsupervised2021} have contributed significantly to improving performance on underrepresented classes. However, a common limitation across these methods is the neglect of inter-class relationships, particularly how semantic similarity between classes influences misclassification patterns.

Our empirical analysis reveals that misclassifications of minority classes often occur in favor of semantically similar majority classes.
To capture and leverage these subtle relationships, we introduce the Relation Contextual Module (\rcm). Unlike general class imbalance metrics that can be inferred from data distributions, inter-class dynamics must be learned from the model's behavior during training. \rcm facilitates this by generating a confusion matrix at each training batch, comparing the model’s predictions with ground-truth labels. This matrix is normalized with respect to the ground truth, enabling a dynamic estimation of misclassification biases between specific class pairs.
To ensure stability and robustness over time, we apply an Exponential Moving Average (\ema) to iteratively update a global confusion matrix. This evolving matrix serves as a reliable representation of the model’s long-term class bias patterns. Beyond providing smoother updates, the use of \ema eliminates the requirement for all classes to be present in every batch—thus streamlining the learning process. The procedure for updating this matrix is detailed in Algorithm~\ref{algo:rcm_update}.

When applied to target domain samples, \rcm relies on high-confidence pseudo-labels to estimate class biases in the absence of ground-truth annotations, offering a reliable approximation that mirrors the supervised setup on the source domain.

%


Aligning source and target domains is common in traditional domain adaptation tasks~\cite{DBLP:conf/cvpr/Kang0YH19,DBLP:conf/cvpr/ZhuPYSL19,DBLP:conf/cvpr/0007CGV19}. While alignment can be performed in either data space~\cite{DBLP:conf/cvpr/ChenZD0D20} or feature space~\cite{DBLP:conf/cvpr/SaitoUHS19}, the absence of source data presents a unique challenge for domain alignment.

Even though the source data is inaccessible, the source-pretrained model retains crucial knowledge about the source domain. Following~\cite{chu2023adversarial}, we partition the target dataset into two groups, leveraging the pre-trained model to establish an explicit source-target domain distinction. To achieve this, a detection variance-based criterion is used, where variances are computed from the pre-trained model’s predictions on the target samples. A higher variance suggests a stronger resemblance to the source domain~\cite{chu2023adversarial}. In particular, the model exhibits greater uncertainty (hard samples) when predicting source-similar images, leading to elevated variance values compared to source-dissimilar images. The detection variance is determined using the following formulation:

\begin{equation}
    v_i = \mathbb{E}[(F_{\theta_s}(X_i) - \mathbb{E}[F_{\theta_s}(X_i)])^2],
\end{equation}
where $F_{\theta_s}(X_i)$ are the predictions of image $X_i$ via the source-pretrained model. Since this calculation is computationally intractable, we approximate it with Monte-Carlo sampling using dropout, following the method proposed by Gal and Ghahramani~\cite{DBLP:conf/icml/GalG16}. This approximation is achieved by conducting $M$ stochastic forward passes while keeping the detection model unchanged~\cite{DBLP:journals/corr/BlundellCKW15}.

Since the outputs \( F_{\theta_s}(X_i) = (\mathbf{b_i}, \mathbf{c_i}) \) consist of bounding box coordinates and class scores, the detection variance is defined as the product of the bounding box variance \( v_{bi} \) and the class score variance \( v_{ci} \). For a given prediction with \( N_i \) bounding boxes and \( K \) classes, where \( \mathbf{\{b_{ij}} = (x_{ij}^1, y_{ij}^1, x_{ij}^2, y_{ij}^2)\}_{j=1}^{N_i} \) and \( \mathbf{\{c_{ij}} = (c_{ij}^1, c_{ij}^2, \ldots, c_{ij}^K)\}_{j=1}^{N_i} \), we can express \( v_{bi} \) and \( v_{ci} \) as follows:

\begin{equation}
    v_{bi} = \frac{1}{M N_i} \sum_{j=1}^{N_i} \sum_{m=1}^{M} \| \mathbf{b_{ij}^m} - \mathbf{\bar{b}_{ij}} \|^2,
\end{equation}
\begin{equation}
    v_{ci} = \frac{1}{M N_i} \sum_{j=1}^{N_i} \sum_{m=1}^{M} \| \mathbf{c_{ij}^m} - \mathbf{\bar{c}_{ij}} \|^2,
\end{equation}
where \( \mathbf{b_{ij}^m} \) and \( \mathbf{c_{ij}^m} \) represent the localization coordinates and classification scores of the \( m \)-th forward pass for the \( j \)-th bounding box in \( X_i \), respectively, and \( \mathbf{\bar{b}_{ij}} \), \( \mathbf{\bar{c}_{ij}} \) denote their corresponding average values over all \( M \) forward passes.

The detection variance for an image \( X_i \) is computed as \( v_i = v_{bi} v_{ci} \). We then rank the images based on their variances, where \( r_i \) represents the rank of \( X_i \). The variance level \( vl_i \) for the \( i \)-th image is given by \( vl_i = \frac{r_i}{N} \). If \( vl_i \geq \sigma \), we categorize \( X_i \) as source-similar; otherwise, it is deemed source-dissimilar, with \( \sigma \in (0, 1) \) being a predefined threshold. This process effectively partitions the target domain into source-similar and source-dissimilar subsets for query-based adversarial alignment between domains.

Oversampling is a widely used strategy in classification tasks to mitigate class imbalance by increasing the representation of minority classes. However, this method presents challenges in object detection, where individual images often contain multiple object classes. Our analysis of breast cancer detection datasets such as DDSM~\cite{lee2017curated} reveals that most images do not consistently feature even a single instance of majority classes, making conventional image-level resampling approaches largely ineffective. This calls for more targeted, instance-aware augmentation techniques tailored to object detection.

Following the instance-level augmentation strategy proposed by Zhang et al.~\cite{Zhang_Pan_Wang_2022}, we extract object instances using bounding box annotations and strategically insert them into other images. To guide this process, we leverage our Relation Contextual Module (\rcm) to identify majority and minority classes based on their likelihood of correct classification.

We compute the average diagonal value of the \rcm—representing the mean self-classification probability—using the formula:
\begin{equation}
{\rm \rcm}_{\rm avg} = \frac{1}{C}\sum^C_{c=0}{\rcm}(c,c),
\end{equation}
where $C$ is the number of classes. Classes with ${\rm \rcm}(c,c)$ values above this mean are labeled as majority classes, while those below are considered minority.

Unlike prior methods that randomly overlay instances, we employ a relation-guided MixUp strategy. For each base instance, we select a paired instance through weighted random sampling, where the weights are determined by inter-class misclassification probabilities captured in the \rcm. For majority base instances, we use the column vector ${\rm \rcm}(:,c)$, setting ${\rcm}(c,c) = 0$ to exclude self-augmentation and focus on classes often confused with the majority. For minority base instances, the row vector ${\rm \rcm}(c,:)$ is used without masking, allowing for beneficial self-augmentation.

Paired instances are resized to match the base instance for bounding box consistency. MixUp is then applied as:
\begin{equation}
\begin{aligned}
\hat{I} &= \beta \cdot I_{\rm base} + (1-\beta) \cdot I_{\rm mix},\\
\hat{c} &= \beta \cdot c_{\rm base} + (1-\beta) \cdot c_{\rm mix},
\end{aligned}
\end{equation}
where $I_{\rm base}$ and $I_{\rm mix}$ are the cropped images of the base and mixed instances, and $\hat{I}$ is the resulting augmented image. Correspondingly, $c_{\rm base}$ and $c_{\rm mix}$ denote the respective class vectors, with $\hat{c}$ being the interpolated class label.

To support domain adaptation, we apply this augmentation to both source similar and dissimilar domains. For source-like images, we sample instances from both domains to promote knowledge transfer. For source dissimilar images, we prioritize dissimilar domain instances, falling back to source similar instances only when necessary. Importantly, we refrain from augmenting minority base instances in the target domain to preserve their semantic integrity, helping the model maintain focus on real target distributions and avoid domain drift.

\begin{algorithm}[t]
\caption{Relation Contextual Module (RCM) Update Procedure}
\label{algo:rcm_update}
\begin{algorithmic}[1]
\REQUIRE Global class-relation matrix $\mathbf{R} \in \mathbb{R}^{C \times C}$ initialized to zero; \ema decay factor $\beta \in [0, 1]$.
\WHILE {training}
    \STATE Initialize batch-specific relation matrix $\mathbf{R}_b \in \mathbb{R}^{C \times C}$ to zero.
    \FOR {each training example with ground-truth label $c_i$ and predicted label $x_i$}
        \STATE $\mathbf{R}_b[c_i, x_i] \gets \mathbf{R}_b[c_i, x_i] + 1$
    \ENDFOR
    \FOR {each class $c \in \{1, \dots, C\}$}
        \STATE Normalize $\mathbf{R}_b[c, :]$ to sum to 1.
        \STATE Update $\mathbf{R}[c, :] \gets \beta \cdot \mathbf{R}[c, :] + (1 - \beta) \cdot \mathbf{R}_b[c, :]$
    \ENDFOR
\ENDWHILE
\end{algorithmic}
\end{algorithm}

To enable inter-semantic Augmentation, we maintain a dynamic repository of class-specific instance crops, referred to as the \textit{Cropbank}~\cite{Zhang_Pan_Wang_2022}. These instance crops are extracted from bounding box annotations in labeled source images as well as pseudo-labeled target images. To facilitate domain-specific augmentation, separate Cropbanks are maintained for source-similar and source-dissimilar subsets. Each Cropbank is updated in a first-in-first-out (\fifo) manner to ensure sample diversity and freshness. This strategy is particularly advantageous for the source dissimilar Cropbank, where early-stage pseudo-labels may be less reliable; replacing them over time helps improve augmentation quality as training progresses.

\subsection{Semantic-Aware Loss}
To further mitigate class bias, we introduce a weighted parameter into the classification loss, guided by the Relation Contextual Module (\rcm) for foreground classes. This weighting scheme emphasizes classes that are frequently misclassified as the dominant majority classes, thereby directing the model’s attention toward improving performance on underrepresented or confused categories.

To accentuate this focus, we apply a non-linear transformation to the \rcm values as follows:
\begin{equation}
    w_i= 
\begin{cases}
    \sqrt{1 - \text{\rcm}(c_i, x_i)}, & \text{if } c_i = x_i \\
    \sqrt{\frac{\text{\rcm}(c_i, x_i)}{\text{\rcm}(c_i, c_i)}}, & \text{otherwise}
\end{cases}
\label{eq:icl_weight}
\end{equation}
where $w_i$ denotes the weight for the $i^{\text{th}}$ instance, $c_i$ is its pseudo-label, and $x_i$ is the predicted class. When $c_i \neq x_i$, we normalize using the diagonal entry $\text{\rcm}(c_i, c_i)$ to scale the misclassification likelihood relative to the class's overall performance.

Background class weights are uniformly set to 1 to prevent model bias against them. To reconcile the disparity between foreground and background weights, we normalize foreground instance weights so that their mean equals that of the background class:
\begin{equation}
W_f = \frac{W_f}{\text{mean}(W_f)},
\end{equation}
where $W_f$ represents the set of weights for foreground instances.

To avoid excessive influence from extreme weights, we apply a regularization term $\lambda_l$ to all class-relation weights:
\begin{equation}
W = \frac{W + \lambda_l}{1 + \lambda_l},
\label{eq:regularization}
\end{equation}
ensuring smoother gradients and preventing the loss from being overly sensitive to outlier weights.

The weighted classification loss is then defined as:
\begin{equation}
\mathcal{L}_{\text{cls}} = \frac{1}{N} \sum_{i=1}^{N} w_i \cdot \text{\ce}(c_i, x_i),
\label{eq:semanticloss}
\end{equation}
where $N$ is the total number of instances, and $\text{\ce}(\cdot)$ denotes the standard cross-entropy loss.

\subsection{Grounded Supervision}

We introduce a framework that leverages Large Vision Foundation Models (\lvfms) through expert distillation, as depicted in Figure~\ref{main_arch}. We introduce an Expert branch aimed at assimilating the strengths of \lvfms into a unified learning approach. As an initial assumption, we posit that the expert model can effectively represent a wide variety of images sourced from datasets like \texttt{ImageNet1K} or \texttt{ImageNet21k} \cite{deng2009imagenet}, \laion400m \cite{schuhmann2021laion}, or \texttt{DataComp1B} \cite{gadre2024datacomp}. We have considered many prominent families of teacher models: \clip \cite{radford2021learning}, \texttt{DINOV2} \cite{oquab2023dinov2}, \vit~\cite{dosovitskiy2020image} and \sam \cite{kirillov2023segment}, \groundingdino~\cite{liu2024groundingdinomarryingdino}, and \biomedparse~\cite{Zhao_2024}. These models have exhibited outstanding performance across a wide array of tasks, with \clip demonstrating broad proficiency, \texttt{DINOV2} excelling in downstream dense tasks such as semantic segmentation under linear probe, and \sam particularly shining in open-vocabulary segmentation tasks.
The newly introduced Expert branch incorporates the pretrained \lvfms. In the context of knowledge distillation, utilizing a single expert's embedding directly for distilling knowledge to the student model may not suffice. We found that relying on an expert's embedding directly for distilling knowledge to the student model is insufficient for effective pseudo-label correction during adaptation. Hence, we propose using expert predictions and introducing a pseudo-supervised loss to guide the student model out from the local minima.

Our objective is to transfer the expertise of the expert model to the student model. The expert model $E(\cdot)$ is frozen, while the student model $S(\cdot)$ is updated via gradient descent during adaptation.
To supervise the student using pseudo labels from the expert, we employ a combination of a consistency loss (mean teacher loss) and a bounding box regression loss.

Let $x_i$ be an input image. The expert produces pseudo-labels in the form of bounding box predictions $E_{\text{bbox}}^i = E(x_i)$, which are used to supervise the student’s prediction $S_{\text{bbox}}^i = S(x_i)$. We define the loss as:
\begin{align}
\mathcal{L}_{\text{expert}} =\ & \lambda_{\text{cls}} \cdot \mathcal{L}_{\text{cls}}(S_{\text{bbox}}^i, E_{\text{bbox}}^i) \notag \\
& + \lambda_{\text{reg}} \cdot \mathcal{L}_{\text{reg}}(S_{\text{bbox}}^i, E_{\text{bbox}}^i)
\label{eq_expert}
\end{align}
where $\mathcal{L}_{\text{cls}}$ is the classification loss (semantic aware loss), and $\mathcal{L}_{\text{reg}}$ is the bounding box regression loss (e.g., smooth \texttt{L1} or \texttt{GIoU} loss). The weights $\lambda_{\text{cls}}$ and $\lambda_{\text{reg}}$ control the contribution of each term.

This unified loss ensures that the student model learns both semantic consistency and localization accuracy from the expert model via pseudo-supervision.

\subsection{Overall Objective and Training Strategy.}

The final objective combines supervised, unsupervised, and discriminator losses:
\begin{equation}
\mathcal{L} = \lambda_u \mathcal{L}_{\text{stu}} + \lambda_d \mathcal{L}_{\text{dis}} + \mathcal{L}_{\text{Expert}},
\end{equation}
where $\lambda_u$ \text{and} $\lambda_d$ control the contributions of the unsupervised and discriminator losses, respectively, and $\mathcal{L}_{\text{stu}},\ \text{and}\ \mathcal{L}_{\text{Expert}}$  have been formulated in Eq.  \eqref{eq1}, and \eqref{eq_expert} respectively. As the teaching process continues, an adequate amount of pseudo-boxes will be produced, and the influence of $\lambda_{u}$ should be lowered to avoid the encoder over-fitting to pseudo-labels. As a result, We decay $\lambda_{d}$ as adaptation continues. More details are discussed in the Supplementary material.

\section{Experiments}
\label{experiments}

We conduct extensive experiments to assess the effectiveness of our approach across various challenging medical imaging datasets, as well as standard \uda and \sfod benchmarks. In the \uda setting, both source and target domain data are accessible during training, facilitating adaptation. Conversely, in the \sfod scenario, adaptation is performed using only a source-trained model without access to source domain data. Additionally, we perform ablation studies employing diverse exchange strategies to validate the efficacy of our proposed methods. We further analyze the promising results obtained by our method through comprehensive visualizations and component-wise evaluations.

\begin{table*}[t]
\centering
  \resizebox{\linewidth}{!}{
    \begin{tabular}{c|l|l|c|cccc|cc}
    \toprule
    \textbf{Exp}& \textbf{Method} & \textbf{Venue} & \textbf{SF} &\textbf{R@0.05}  & \textbf{R@0.3} &\textbf{R@0.5}& \textbf{R@1.0}  & \textbf{AUC} & \textbf{F1-score} \\ 
    \midrule
    & \umt~\cite{10.1007/978-3-031-20077-9_37} & CVPR'21 &\color{red}\color{red}\xmark& 0.01 &0.09 &0.15 &0.21 &0.204&0.319\\
    & \sfa~\cite{wang2021exploring} & MM'21 & \color{red}\color{red}\xmark & 0.03 &0.13 &0.23 &0.35 &0.241&0.338\\

    & \h2fa~\cite{xu2022h2fa} & CVPR'22 &\color{blue}\color{red}\xmark& 0.13 &0.32 &0.45&0.52 &0.575&0.312\\
    & \aqt~\cite{huang2022aqt} & IJCAI'22 &\color{blue}\color{red}\xmark&0.11 &0.37 &0.44&0.57&0.680& 0.349\\
    & \at~\cite{li2022cross} & CVPR'22 & \color{red}\color{red}\xmark & 0.19 &0.38 &0.65 &0.75 &0.721&0.512\\
DDSM & \dadapt~\cite{jiang2022decoupled} & ICLR'22 &\color{red}\color{red}\xmark& 0.11 &0.29 &0.45 &0.59 &0.731&0.414\\

    to& \HT~\cite{Deng_2023_CVPR} & CVPR'23 & \color{blue}\color{red}\xmark & 0.17 & 0.49 & 0.61 & 0.69 & 0.704 & 0.362 \\
    
INBreast& \clipgap~\cite{vidit2023clip} & CVPR'23 & \color{blue}\color{red}\xmark & 0.15 & 0.55 & 0.61 & 0.75 & 0.712 & 0.445 \\
    & \confmix~\cite{mattolin2023confmix} & WACV'23 & \color{blue}\color{red}\xmark & 0.19 & 0.47 & 0.58 & 0.73 & 0.737 & 0.409 \\
    & \mrt~\cite{10376698} & ICCV'23 & \color{blue}\color{red}\xmark & 0.16 & 0.54 & 0.64 & 0.72 & 0.789 & 0.489 \\

    & \dmaster~\cite{ashraf2024dmastermaskannealedtransformer}  & MICCAI'24 &\color{red}\xmark&0.25 &0.61 &0.70&0.82&0.808& 0.524\\

    & \mexformer~\cite{wang2021exploring} & MM'21 &\color{blue}\checkmark& 0.02 &0.03 &0.03&0.03 &0.060&0.090\\
    & \irg~\cite{deng2021unbiased} & CVPR'23 &\color{blue}\checkmark&\underline{0.05} &0.05 &0.07&0.09&0.110& 0.120\\
    & \lpld~\cite{yoon2024enhancing} & ECCV'24 &\color{blue}\checkmark & \textbf{0.09} & \underline{0.15} & \underline{0.35} & \underline{0.35} & \underline{0.548} & \underline{0.635} \\
    \midrule
    \rowcolor[gray]{0.87}
    & \gt (Ours) &  &\color{blue}\checkmark & \underline{0.06} & \textbf{0.45} & \textbf{0.65} & \textbf{0.92} & \textbf{0.589}&\textbf{0.758}\\
    \bottomrule
    \end{tabular}
      }

\caption{ Results of adaptation from DDSM to INBreast. Bold values indicate the best performance, while underlined values indicate the second-best performance.}

 \label{tab:d2I}
\end{table*}

\subsection{Datasets}

The following datasets were used in our experiments.
\begin{enumerate}
    \item \textbf{DDSM}~\cite{lee2017curated} A publicly available \bcd dataset comprising 2,620 full mammography images with 1,162 malignancies. For full details, please refer to the Supplementary Material.
    \item \textbf{INBreast}~\cite{moreira2012inbreast} A smaller \bcd dataset with 410 mammography images from 115 patients, including 87 malignancies.
    \item \textbf{RSNA}~\cite{carr2022rsna} The original RSNA dataset consists of 54,706 screening mammograms containing 1,000 malignancies from 8,000 patients. We curated a subset named RSNA-BSD1K, comprising 1,000 mammograms with 200 malignant cases annotated at the bounding box level by expert radiologists.
    \item \textbf{Cityscapes}~\cite{cityscapes} Collected from urban scenes containing 2,975 training and 500 validation images.
    \item \textbf{Foggy Cityscapes}~\cite{foggycityscapes} Constructed using a fog synthesis algorithm from Cityscapes. We use Cityscapes as the source domain and Foggy Cityscapes (fog density = 0.02) as the target domain. Similar to Cityscapes, this dataset integrates fog and depth information into street-view imagery.
    \item \textbf{BDD100k}~\cite{bdd100k} Created at UC Berkeley, this dataset contains 100K annotated 720p images (1 frame per video) across 10 classes from diverse US driving scenes, supporting robust detection and adaptation studies.
   \item \textbf{InSeg1}~\cite{9183116} A large-scale surgical segmentation dataset containing endoscopic video frames annotated at the pixel level across multiple surgical tools and anatomical structures. We extract bounding boxes from segmentation masks to enable detection-style training.
    \item \textbf{InSeg2}~\cite{9183116} A companion dataset to InSeg1 featuring different surgical environments, camera viewpoints, and lighting conditions. It is used as the target domain in the InSeg1$\rightarrow$InSeg2 transfer setting to evaluate robustness under real-world surgical domain shifts.

\end{enumerate}

\begin{table*}[t]
\centering
  \resizebox{\linewidth}{!}{
\begin{tabular}{c|l|l|c|cccc|cc}

    \toprule
    \textbf{Exp}& \textbf{Method} & \textbf{Venue} & \textbf{SF} &\textbf{R@0.05}  & \textbf{R@0.3} &\textbf{R@0.5}& \textbf{R@1.0}  & \textbf{AUC} & \textbf{F1-score} \\ 
    \midrule
    & \dadapt~\cite{jiang2022decoupled} & ICLR'21 & \color{red}\xmark & 0.04 &0.12 &0.18 &0.29 &0.439&0.263\\
    & \at~\cite{li2022crossdomain} & CVPR'22 &\color{red}\xmark& 0.16 &0.28 &0.35 &0.42 &0.486&0.338\\
    DDSM& \h2fa~\cite{9878659} & CVPR'22 &\color{red}\xmark& 0.03 &0.13 &0.18 &0.36 &0.634&0.236\\
    to & \mrt~\cite{zhao2023masked} & ICCV'23 & \color{red}\xmark & 0.32 &0.52 &0.69 &0.72 &0.741&0.352\\
    RSNA& \mexformer~\cite{wang2021exploring} & MM'21 &\color{blue}\checkmark& \textbf{0.24} &\underline{0.31} &\underline{0.39}&\underline{0.39} &\underline{0.336}&\underline{0.287}\\
    & \irg~\cite{deng2021unbiased} & CVPR'23 &\color{blue}\checkmark&\underline{0.16} &0.25 &0.37&\underline{0.39}&0.308& 0.235\\
    \midrule
    \rowcolor[gray]{0.87}
    & \gt (Ours)  &  &\color{blue}\checkmark& 0.10 &\textbf{0.43} &\textbf{0.58}&\textbf{0.91}&\textbf{0.781}& \textbf{0.530}\\
    \bottomrule
    \end{tabular}
      }
\caption{Results of adaptation from DDSM to RSNA. Bold values indicate the best performance, while underlined values indicate the second-best performance.}
  \label{tab:med2}
\end{table*}
\begin{table*}[t]
\centering
  \resizebox{\linewidth}{!}{
    \begin{tabular}{c|l|l|c|cccc|cc}

    \toprule
    \textbf{Exp}& \textbf{Method} & \textbf{Venue} & \textbf{SF} &\textbf{R@0.05}  & \textbf{R@0.3} &\textbf{R@0.5}& \textbf{R@1.0}  & \textbf{AUC} & \textbf{F1-score} \\ 
    \midrule
    & \dadapt~\cite{jiang2022decoupled} & ICLR'21 & \color{red}\xmark & 0.00 &0.06 &0.09 &0.10 &0.381&0.362\\
    & \at~\cite{li2022crossdomain} & CVPR'22 &\color{red}\xmark& 0.01 &0.08 &0.10 &0.15 &0.385&0.311\\
    RSNA& \h2fa~\cite{9878659} & CVPR'22 &\color{red}\xmark& 0.02 &0.08 &0.10 &0.12 &0.483&0.315\\
    to & \mrt~\cite{zhao2023masked} & ICCV'23 &\color{red}\xmark & 0.03 & 0.09 & 0.12 & 0.17 &0.739&0.587\\
    INBreast& \mexformer~\cite{wang2021exploring} & MM'21 &\color{blue}\checkmark& \underline{0.02} &0.03 &0.03&0.03 &0.060&0.090\\
    & \irg~\cite{deng2021unbiased} & CVPR'23 &\color{blue}\checkmark&\textbf{0.05} &\underline{0.05} &\underline{0.07}&\underline{0.09}&\underline{0.110}& \underline{0.120}\\
    \midrule
    \rowcolor[gray]{0.87}
    & \gt (Ours) &  &\color{blue}\checkmark& 0.01 &\textbf{0.28} &\textbf{0.49}&\textbf{0.90}&\textbf{0.638}& \textbf{0.605}\\
    \bottomrule
    \end{tabular}
      }
\caption{Results of adaptation from RSNA to INBreast. Bold values indicate the best performance, while underlined values indicate the second-best performance.}
  \label{tab:med3}
\end{table*}

\begin{table*}[t]
\centering
\resizebox{\textwidth}{!}{%
\begin{tabular}{l|l|c|cccccccc|c}

\toprule
\textbf{Method} & \textbf{Venue} & \textbf{SF} & \textbf{Person} & \textbf{Rider} & \textbf{Car} & \textbf{Truck} & \textbf{Bus} & \textbf{Train} & \textbf{Mcycle} & \textbf{Bicycle} & \textbf{mAP} \\
\midrule
\rowcolor[gray]{0.96}
\texttt{Source Only} & &\color{red}\xmark & 22.4 & 26.6 & 28.5 & 9.0  & 16.0 & 4.3 & 15.2 & 25.3 & 18.4 \\
\texttt{DA-Faster}~\cite{Chen_2018_CVPR}  & CVPR'18 & \color{red}\xmark & 29.2 & 40.4 & 43.4 & 19.7 & 38.3 & 28.5 & 23.7 & 32.7 & 32.0 \\
\texttt{EPM}~\cite{hsu2020every} & ECCV'20 & \color{red}\xmark& 44.2 & 46.6 & 58.5 & 24.8 & 45.2 & 29.1 & 28.6 & 34.6 & 39.0 \\
\texttt{SSAL}~\cite{munir2021ssal} & NIPS'21 & \color{red}\xmark& 45.1 & 47.4 & 59.4 & 24.5 & 50.0 & 25.7 & 26.0 & 38.7 & 39.6 \\
\texttt{SFA}~\cite{wang2021exploring} & MM'21 & \color{red}\xmark&46.5 & 48.6 & 62.6 & 25.1 & 46.2 & 29.4 & 28.3 & 44.0 & 41.3 \\
\texttt{UMT}~\cite{10.1007/978-3-031-20077-9_37} & CVPR'21 &\color{red}\xmark & 33.0 & 46.7 & 48.6 & 34.1 & 56.5 & 46.8 & 30.4 & 37.3 & 41.7 \\
\texttt{D-adapt}~\cite{jiang2022decoupled} & ICLR'21 & \color{red}\xmark & 40.8 & 47.1 & 57.5 & 33.5 & 46.9 & 41.4 & 33.6 & 43.0 & 43.0 \\

\texttt{AT}~\cite{li2022cross} & CVPR'22 &\color{red}\xmark &43.7 & 54.1 & 62.3 & 31.9 & 54.4 & 49.3 & 35.2 & 47.9 & 47.4 \\

\texttt{CSDA}~\cite{gao2023csda} & ICCV'23 & \color{red}\xmark & 46.6 & 46.3 & 63.1 & 28.1 & 56.3 & 53.7 & 33.1 & 39.1 & 45.8 \\
\texttt{HT}~\cite{deng2023harmonious} & CVPR'23 & \color{red}\xmark & 52.1 & 55.8 & 67.5 & 32.7 & 55.9 & 49.1 & 40.1 & 50.3 & 50.4 \\
\texttt{MRT}~\cite{zhao2023masked} & ICCV'23 & \color{red}\xmark& 52.8 & 51.7 & 68.7 & 35.9 & 58.1 & 54.5 & 41.0 & 47.1 & 51.2 \\

\texttt{SFOD}~\cite{li2020free} & AAAI'21 & \color{blue}\checkmark & 21.7 & 44.0 & 40.4 & 32.6 & 11.8 & 25.3 & 34.5 & 34.3 & 30.6 \\
\texttt{SFOD-Mosaic}~\cite{li2020free} & AAAI'21 & \color{blue}\checkmark & 25.5 & 44.5 & 40.7 & 33.2 & 22.2 & 28.4 & 34.1 & 39.0 & 33.5 \\
\texttt{HCL}~\cite{huang2022model} & NIPS'21 & \color{blue}\checkmark & 38.7 & 46.0 & 47.9 & 33.0 & 45.7 & 38.9 & 32.8 & 34.9 & 39.7 \\
\texttt{SOAP}~\cite{soapijis21} & IJIS'21 & \color{blue}\checkmark & 35.9 & 45.0 & 48.4 & 23.9 & 37.2 & 24.3 & 31.8 & 37.9 & 35.5 \\
\texttt{LODS}~\cite{li2022source} & CVPR'22 & \color{blue}\checkmark & 34.0 & 45.7 & 48.8 & 27.3 & 39.7 & 19.6 & 33.2 & 37.8 & 35.8 \\
\texttt{AASFOD}~\cite{chu2023adversarial} & AAAI'23 & \color{blue}\checkmark & 32.3 & 44.1 & 44.6 &28.1 & 34.3 & 29.0 & 31.8 & 38.9 & 35.4 \\
\texttt{IRG}~\cite{vs2023instance} & CVPR'23& \color{blue}\checkmark & 37.4 & 45.2 & 51.9 & 24.4 & 39.6 & 25.2 & 31.5 & 41.6 & 37.1 \\
\texttt{PETS}~\cite{liu2023periodically} & ICCV'23 & \color{blue}\checkmark & \textbf{46.1} & \underline{52.8} & \textbf{63.4} & 21.8 & 46.7 & 5.5 & \underline{37.4} & \underline{48.4} & 40.3 \\
\texttt{DACA}~\cite{10.1007/s11263-024-02170-z} & IJCV'24 & \color{blue}\checkmark & \underline{44.7} & 31.2 & 60.1 & \textbf{53.1} & \underline{53.9} & 0.0 & 27.8 & 45.9 & 39.9 \\
\texttt{LPLD}~\cite{yoon2024enhancing} & ECCV'24 & \color{blue}\checkmark & 39.7 & 49.1 & 56.6 & 29.6 & 46.3 & 26.4 & 36.1 & 43.6 & 40.9 \\
\texttt{SF-UT}~\cite{hao2024simplifying} & ECCV'24 & \color{blue}\checkmark & 40.9 & 48.0 & 58.9 & 29.6 & 51.9 & \underline{50.2} & 36.2 & 44.1 & \underline{45.0} \\
\midrule
\rowcolor[gray]{0.87}
\gt (Ours) &  & \color{blue}\checkmark & 42.7 & \textbf{55.4}  & \underline{61.7} & \underline{40.7} & \textbf{62.0} & \textbf{54.6} & \textbf{39.1} & \textbf{53.0} &  \textbf{50.8} \\
\midrule

\texttt{Oracle} &  & \color{red}\xmark &66.3 & 61.1 & 80.8 & 45.6 & 68.8 & 52.0 & 49.1 & 54.9 & 59.8  \\
\bottomrule
\end{tabular}%
}
\caption{ Results of adaptation from Normal Cityscapes to Foggy weather Cityscapes (\cf). “\texttt{SF}” refers to source-free setting. “\texttt{Oracle}” refers to the models trained by using labels during training. Bold values indicate the best performance, while underlined values indicate the second-best performance.}
\label{c2f}

\end{table*}

\subsection{Implementation Details}
Building upon existing research~\cite{huang2021model, li2021free, zhao2023masked, oza2023unsupervised, chu2023adversarial, liu2023periodically}, we validate our method across state-of-the-art \sfod and \uda benchmarks. We utilize the Faster R-CNN~\cite{frcnn} architecture with VGG-16 and ResNet-101 as backbone networks. The model's hyperparameters are set as follows: an Exponential Moving Average (\ema) decay rate of 0.9996, beta-distribution parameters of $[0.5, 0.5]$, an adversarial loss weight ($\lambda_d$) of 0.1, an unsupervised loss weight ($\lambda_u$) of 1.0, and a regularization term ($\lambda_l$) of 1.0. A hard threshold ($\tau$) of 0.8 is applied for pseudo-labeling. We implement weak-strong augmentation on both source and target images. 
The training process involves initially training the student model for 20{,}000 iterations using labeled source data. Subsequently, the teacher model is initialized with the student's parameters and updated through \ema at each iteration. Training continues for an additional 60{,}000 iterations, incorporating both labeled source and unlabeled target data. All experiments were performed on a  2 NVIDIA H100 (80 GB) GPUs under identical settings as the baseline. Adding RCM + SAL increases training time by only $\sim$2–3 \% with $<$ 0.3 GB extra VRAM, while the training-only Expert branch adds $\sim$10–12 \% time and $\sim$1 GB VRAM. These modules are disabled during inference, keeping latency and FLOPs identical to the baseline. Profiling details are provided in Appendix
Our framework is built upon the Detectron2~\cite{detectron2} platform. Experiments are conducted with a batch size of 8 source and 8 target images, distributed across 2 NVIDIA H100 GPUs. Further experimental setup details are provided in the Appendix.

\subsection{Evaluation Metrics}
For Natural Images, we report the \ap for each class and \map score, following previous works. For Medical Images, we use the Free-Response Receiver Operating Characteristic
(\froc) curves for detection, F1-score, and \auc for reporting our classification results. The curves provide a graphical representation of sensitivity/recall values at different False Positives per Image (\fpi). We follow related works in this area~\cite{rangarajan2023deep} and consider a prediction as true positive if the center of the predicted bounding box lies within the ground truth
box.

\subsection{Comparison with Current \sota Methods}

We evaluate the performance of our proposed Grounded Teacher (\gt) method against other approaches on all three medical benchmarks and the natural benchmark mentioned earlier for generalizability. Since \uda and \sfdaod share similar task settings, we conducted comparisons with both. Table \ref{tab:d2I}, Table~\ref{tab:med2} and Table~\ref{tab:med3} present the comparison results on medical image datasets. Table~\ref{c2f} presents a comparison of the natural dataset. Our proposed \gt consistently outperforms existing all \sota methods, demonstrating generalizability and significant improvements across both natural and medical settings.

\noindent \textbf{DDSM to INBreast.}
Adaptation from large to small-scale medical datasets with different modalities. Here, we consider DDSM~\cite{lee2017curated} dataset as the source domain and {INBreast} \cite{moreira2012inbreast} as the target domain. Results are presented in Table~\ref{tab:d2I}. Our proposed method demonstrates superior performance across various False Positives per Image (\fpi) values compared to existing methods as displayed.

\noindent \textbf{RSNA to INBreast.}
Adaptation across medical datasets with different machine-acquisitions. This is vital for enhancing healthcare outcomes, improving diagnostic accuracy, and facilitating better clinical decisions. To evaluate our method's performance, we adapt a model trained on DDSM~\cite{lee2017curated} to RSNA~\cite{carr2022rsna}. Results for all \fpi values are presented in Table~\ref{tab:med2}, demonstrating that our method achieves state-of-the-art performance on this benchmark.

\noindent \textbf{DDSM to RSNA.} 
This experiment evaluates domain adaptation across datasets collected from distinct geographical regions. Specifically, we consider the DDSM~\cite{lee2017curated} dataset as the source domain and RSNA~\cite{carr2022rsna} as the target domain. As shown in Table~\ref{tab:med2}, our proposed method consistently outperforms existing approaches across all \fpi thresholds, demonstrating its robustness and effectiveness in cross-domain generalization.
\begin{table}[t]
\centering
\resizebox{\columnwidth}{!}{%
\begin{tabular}{lccccc}
\toprule
\textbf{Method} & \textbf{R@0.05} & \textbf{R@0.3} & \textbf{R@0.5} & \textbf{R@1.0} & \textbf{AUC} \\
\midrule
 Source Only & 0.00 & 0.12 & 0.25 & 0.25 & 0.03 \\
\textbf{GT (Ours)} & \textbf{0.01} & \textbf{0.14} & \textbf{0.28} & \textbf{0.30} & \textbf{0.21} \\
\bottomrule
\end{tabular}%
}
\caption{Results on InSeg1$\rightarrow$InSeg2 dataset. GT consistently improves recall and AUC across thresholds.}
\label{tab:inseg1_to_inseg2}
\color{blue}
\end{table}

\noindent \textbf{Cityscapes to Foggy Cityscapes.}
Object detectors often experience a significant drop in performance when deployed under adverse real-world conditions, such as fog, due to the domain shift caused by the lack of such conditions in the training data. Domain adaptation aims to bridge this gap between normal and adverse weather scenarios. To investigate this, we conduct experiments on the widely-used Cityscapes $\rightarrow$ FoggyCityscapes benchmark.  As presented in Table~\ref{c2f}, student-based frameworks consistently outperform other approaches by a notable margin. 

\noindent\textbf{InSeg1 to InSeg2.} 
The InSeg benchmark features dense endoscopic and surgical imagery with challenging lighting and class overlap. For fair comparison, we extract bounding boxes from segmentation masks as ground-truth annotations. As shown in Table~\ref{tab:inseg1_to_inseg2}, GT improves over the source-only baseline from 0.25$\rightarrow$0.30 at R@1.0 and raises AUC from 0.03$\rightarrow$0.21, demonstrating robust generalization under surgical domain shifts.

\begin{blueblock}

\end{blueblock}


\subsection{Ablation Studies}

\begin{table}[t]
\setlength\tabcolsep{3pt}
\renewcommand{\arraystretch}{1.2}
\centering
\caption{Ablation study on \gt components (Cityscapes$\rightarrow$Foggy). RCM forms the base of all variants.}
\label{tab:ablations}
\begin{tabularx}{\linewidth}{@{}c|*{3}{>{\centering\arraybackslash}X}|>{\centering\arraybackslash}X@{}}
\toprule
\rowcolor{gray!20}
\textbf{Method} & \textbf{RCM} & \textbf{SA} & \textbf{SAL} & \textbf{mAP} \\
\midrule
\rowcolor{gray!4}
Source only (Zero-shot) &  &  &  & 18.4 \\
Base (\at~\cite{at}) &  &  &  & 30.4 \\
\rowcolor{gray!4}
+ \sa & \cmark & \cmark &  & 41.8 \\
\rowcolor{gray!4}
+ \sal & \cmark &  & \cmark & 44.2 \\
\rowcolor{gray!8}
\textbf{\gt (Full)} & \cmark & \cmark & \cmark & \textbf{50.8} \\
\bottomrule
\end{tabularx}
\end{table}

To verify the significance of our contributions, we conducted an ablation study. 
All experiments within this study were performed on the Cityscapes $\to$ Foggy Cityscapes benchmark using the VGG16 backbone.
\subsubsection{Quantitative Ablation}

\begin{figure}[t]
    \centering
    \includegraphics[width=\linewidth]{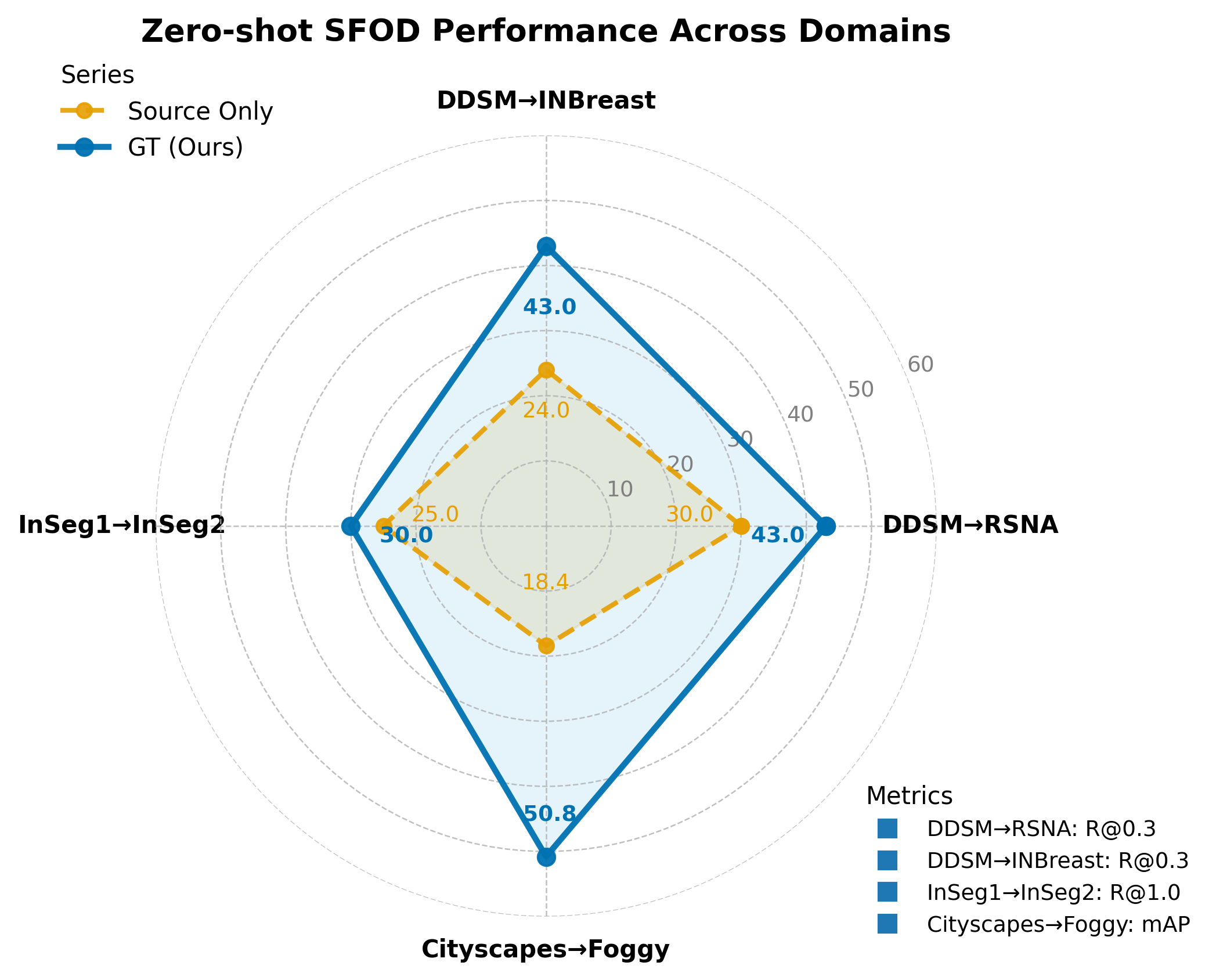}
    \caption{\textbf{Cross-domain zero-shot SFOD performance.} 
    Radar chart summarizing GT's gains across medical (DDSM$\rightarrow$RSNA, DDSM$\rightarrow$INBreast), surgical (InSeg1$\rightarrow$InSeg2), and natural (Cityscapes$\rightarrow$Foggy) transfers. 
    GT consistently surpasses the source-only baseline, demonstrating strong cross-domain generalization.}
    \label{fig:zs_sfod_radar}
\end{figure}

\begin{figure}[t]
    \centering
    \includegraphics[width=\linewidth]{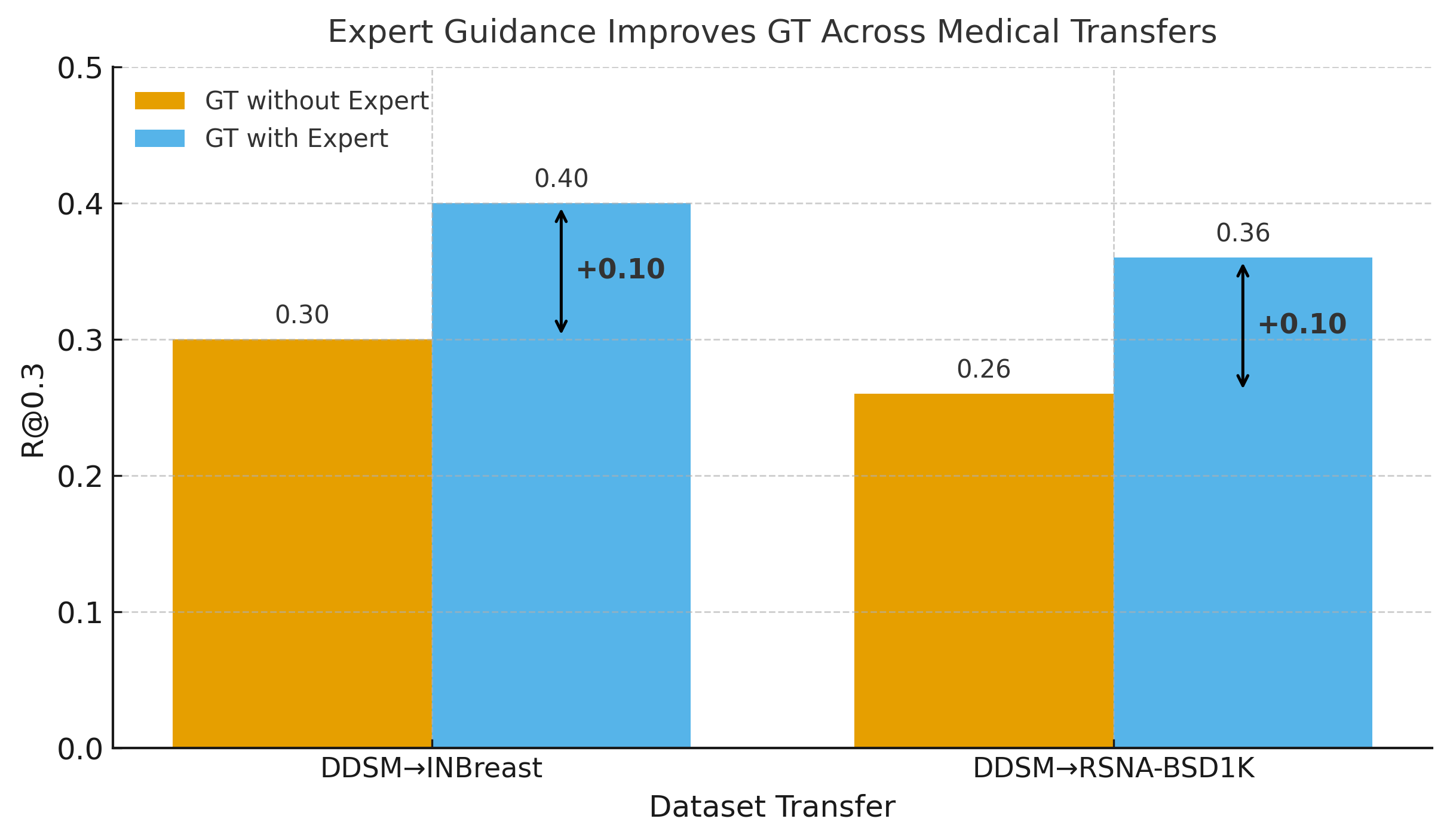}
    \caption{\textbf{Expert-branch ablation.} Visual comparison of GT with and without expert guidance on two medical dataset transfers. The arrows denote absolute gains of +0.10~R@0.3 on both DDSM$\rightarrow$INBreast and DDSM$\rightarrow$RSNA-BSD1K, demonstrating the benefit of expert supervision.}
    \label{fig:placeholder}
\end{figure}

Figure~\ref{fig:zs_sfod_radar} visualizes GT's zero-shot performance across four representative transfers, highlighting consistent improvements across both medical and natural domains. 
The radar layout emphasizes GT’s balanced gains, with particularly large margins in the natural domain (Cityscapes$\rightarrow$Foggy) and stable boosts in surgical and medical transfers.
Table~\ref{tab:ablations} presents a quantitative analysis of the impact of each component within our framework. The base framework prior to integrating our modules aligns with the \at model~\cite{li2022cross}, achieving a 30.4~\map. Given that the Relation-Contextual Module (\rcm) is integral to both class-relation augmentation and semantic-aware loss, it remains a constant across all experimental variations. Integrating Semantic Augmentation (\sa) along with \rcm increases performance to 41.8~\map, a gain of +11.4 over the base. Adding the Semantic-Aware Loss (\sal) with \rcm separately results in a higher \map of 44.2, demonstrating a +13.8 improvement compared to the baseline, indicating that \sal provides a greater performance boost than \sa in this ablation setting. Notably, Class-Relation Augmentation (\cra) significantly reduces the performance disparity between minority and majority classes. Finally, our full \gt model, which combines \rcm, \sa, and \sal, achieves the highest score of 50.8~\map, boosting overall performance and underscoring the method’s efficacy in managing class imbalance.  
In addition, Figure~\ref{fig:placeholder} complements our framework by visualizing the \textbf{Expert-branch ablation results}. It illustrates the improvement in recall (R@0.3) when integrating expert guidance across two medical shifts, showing consistent absolute gains of +0.10. This confirms that the expert branch provides meaningful supervision and stability during adaptation.

\begin{figure*}[t]
\centering
\includegraphics[width=\textwidth]{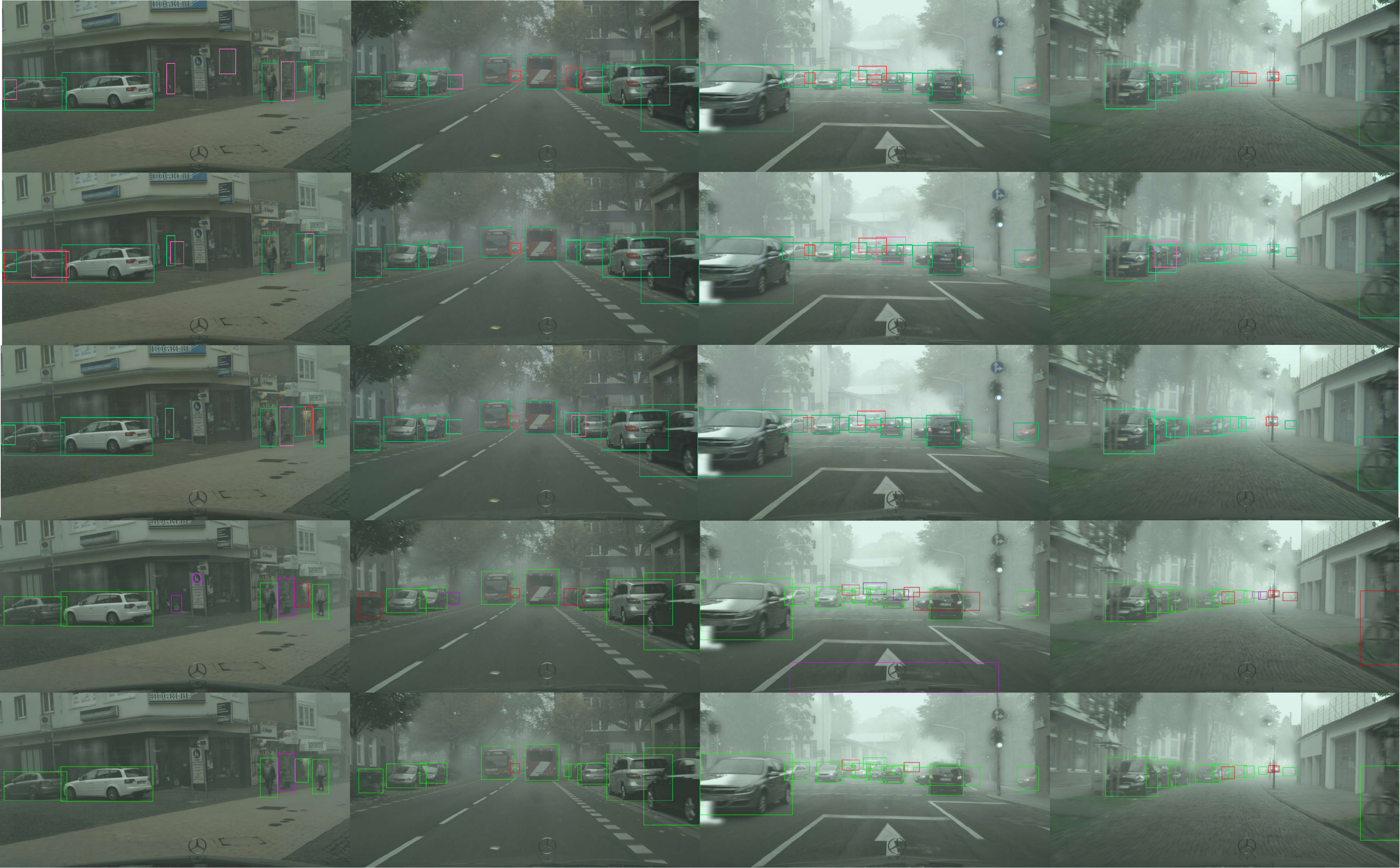}
\caption{Qualitative results of our proposed Grounded Teacher on Natural settings (Cityscapes $\rightarrow$ Foggy Cityscapes). We present visual comparisons between MRT~\cite{10376698} (top row), IRG~\cite{deng2021unbiased}, LPLD~\cite{yoon2024enhancing}, AT~\cite{li2022cross}, GT (bottom row). Our method, GT, effectively mitigates misclassifications, reduces false negatives, and eliminates false positives. Bounding box colors indicate: Green — true positive, Blue — misclassified, Red — false negative, and Pink — false positive.}

\centering
\label{qual_all}
\end{figure*}


\subsubsection{Impact of Augmentation}

In our study, we assess the influence of augmentation strategies, focusing on both the proportion of augmented data and the criteria for selecting instances to augment. Image augmentation plays a pivotal role in our methodology, enhancing the dataset by incorporating additional representations, particularly of minority classes. However, excessive augmentation can lead to the model learning less accurate class representations. To mitigate this, we judiciously augment a random subset of images.
\begin{figure*}[htbp]
    \centering
    \centering
    \includegraphics[width=\linewidth]{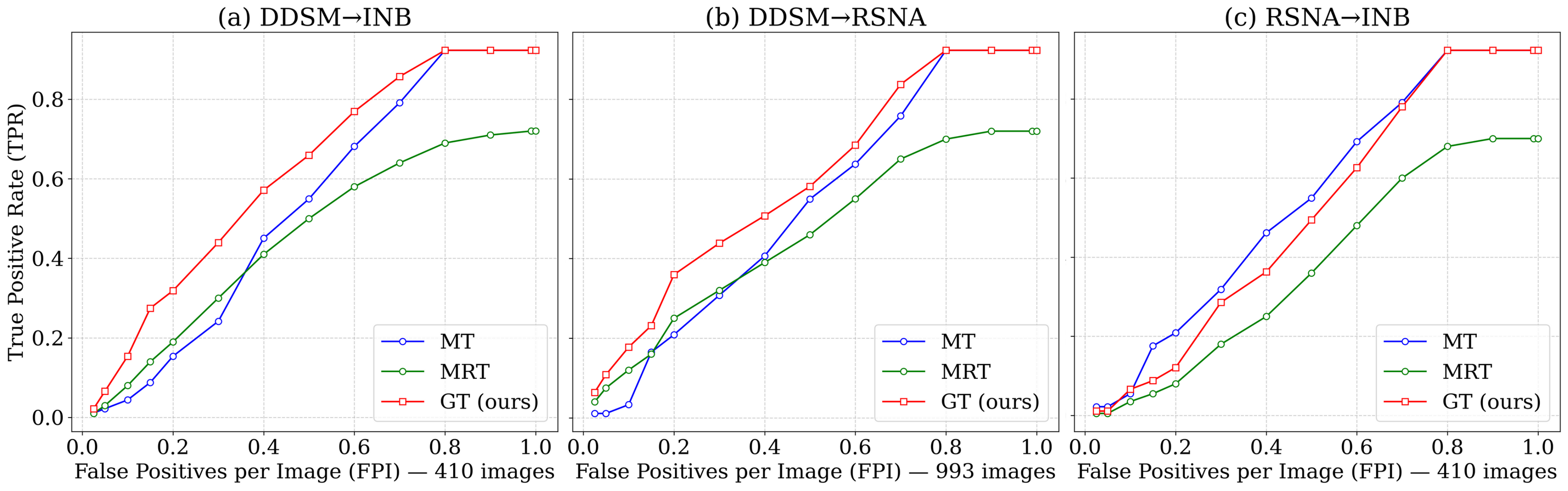}
    \caption{\textbf{FROC analysis of model generalization on unseen target domains}. Performance comparison between our method (GT), the MT baseline, and MRT~\cite{zhao2023masked} for detection tasks under domain shift: (a) DDSM→INBreast (410 images), (b) DDSM→RSNA (993 images), and (c) RSNA→INBreast (410 images), where our method demonstrates superior performance across different domain shifts. Higher curves indicate better trade-offs between sensitivity (\texttt{TPR}) and false positives per image (FPI).}
    \label{fig:froc_plots}
\end{figure*}

The pairing of class instances during augmentation is crucial for improving its effectiveness. Unlike random pairing, our Semantic Augmentation (\sa) method prioritizes pairing instances from closely related minority and majority classes. This targeted approach ensures that the Mixup outputs are more meaningful, thereby enhancing the performance of minority classes.

Table \ref{table:mixup_abla} presents the experimental results of this approach. We compare the effects of randomly selecting class instances for Mixup against our \sa method across different augmentation probabilities, which represent the likelihood of an instance in the base image being augmented. Applying Mixup randomly yields a modest improvement of +1.1 \map in overall performance. In contrast, employing \sa leads to a more substantial increase of +4.4 \map. These findings indicate that pairing highly related classes during Mixup enhances minority class performance with minimal impact on majority classes. Furthermore, our experiments reveal that excessive augmentation can negatively affect model performance, underscoring the importance of a balanced augmentation strategy.
%

\begin{table}[t]
\setlength\tabcolsep{3pt}
\renewcommand{\arraystretch}{1.2}
\centering
\begin{tabularx}{\linewidth}{@{}c|*{2}{>{\centering\arraybackslash}X}@{}}
\toprule
\textbf{Selection Method} & \textbf{mAP}  \\
\midrule
Random (0.5) & 46.4 \\
Semantic Augmentation (1.0)     & 47.5 \\
Semantic Augmentation (0.5)     & 50.8  \\
\bottomrule
\end{tabularx}
\caption{Comparison of selecting class instances randomly and via \cra. Values in brackets refer to the likelihood of an instance being augmented.}
\label{table:mixup_abla}
\end{table}

\subsubsection{Weighing Strategy for Semantic Loss}
To enhance the performance of minority classes, we introduce a Semantic Aware Loss (\sal) function, as defined in Eq~\ref{eq:semanticloss}. Class-level loss functions are a prevalent strategy for addressing dataset class imbalances. We compare our \sal with a variant of existing class-level losses that utilize only the diagonal elements of the Inter-Class Relation module, which correspond to the ground-truth class likelihood for accurate classification. In contrast, \sal leverages the likelihood of the ground-truth being classified as the predicted class to influence its loss function.

As shown in Table \ref{table:weight_alba}, incorporating inter-class relationships through \sal achieves an mAP of 50.8, a performance improvement of +3.7 \map compared to the base. While \sal effectively utilizes inter-class information, there are instances where it might excessively penalize well-performing classes without proper constraints. To mitigate this, we apply a regularization term, as specified in Eq~\ref{eq:regularization}. Omitting this regularization (\sal w/o Reg.) results in an \map of 47.8, demonstrating a significant performance drop of -3.0 \map compared to the full \sal. This clearly highlights the importance of the regularization term, likely preventing certain class weights from becoming disproportionately small during training.

\begin{table}[t]
\setlength\tabcolsep{3pt}
\renewcommand{\arraystretch}{1.2}
\centering
\begin{tabularx}{\linewidth}{@{}c|>{\centering\arraybackslash}X@{}}
\toprule
\textbf{Class Weight Strategy} & \textbf{mAP} \\
\midrule
Class-Level & 47.1 \\
\sal w/o Reg. &47.8 \\
\sal & 50.8 \\
\bottomrule
\end{tabularx}
\caption{Class loss weighting strategies. \textit{Class-Level} uses only the diagonal values in our \rcm with regularization. \textit{\sal} refers to our full Semantic Aware Loss.}
\label{table:weight_alba}
\end{table}

\begin{figure}[t]
    \centering
    \begin{subfigure}{0.48\columnwidth}
        \centering
        \includegraphics[width=0.98\columnwidth]{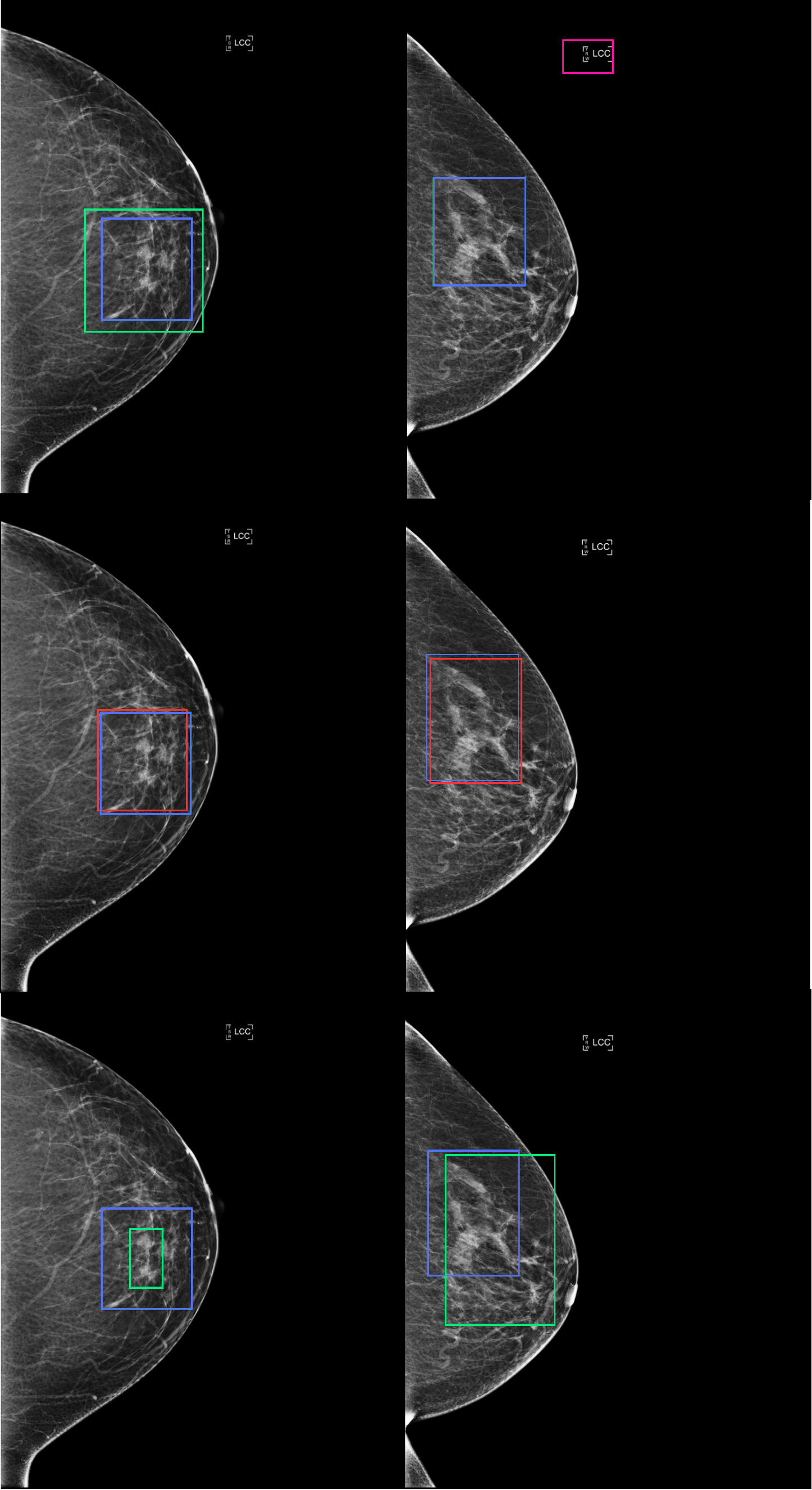}
        \caption{DDSM $\rightarrow$ RSNA}
        \label{fig:ddsm2rsna}
    \end{subfigure}
    \hfill
    \begin{subfigure}{0.45\columnwidth}
        \centering
        \includegraphics[width=0.995\columnwidth]{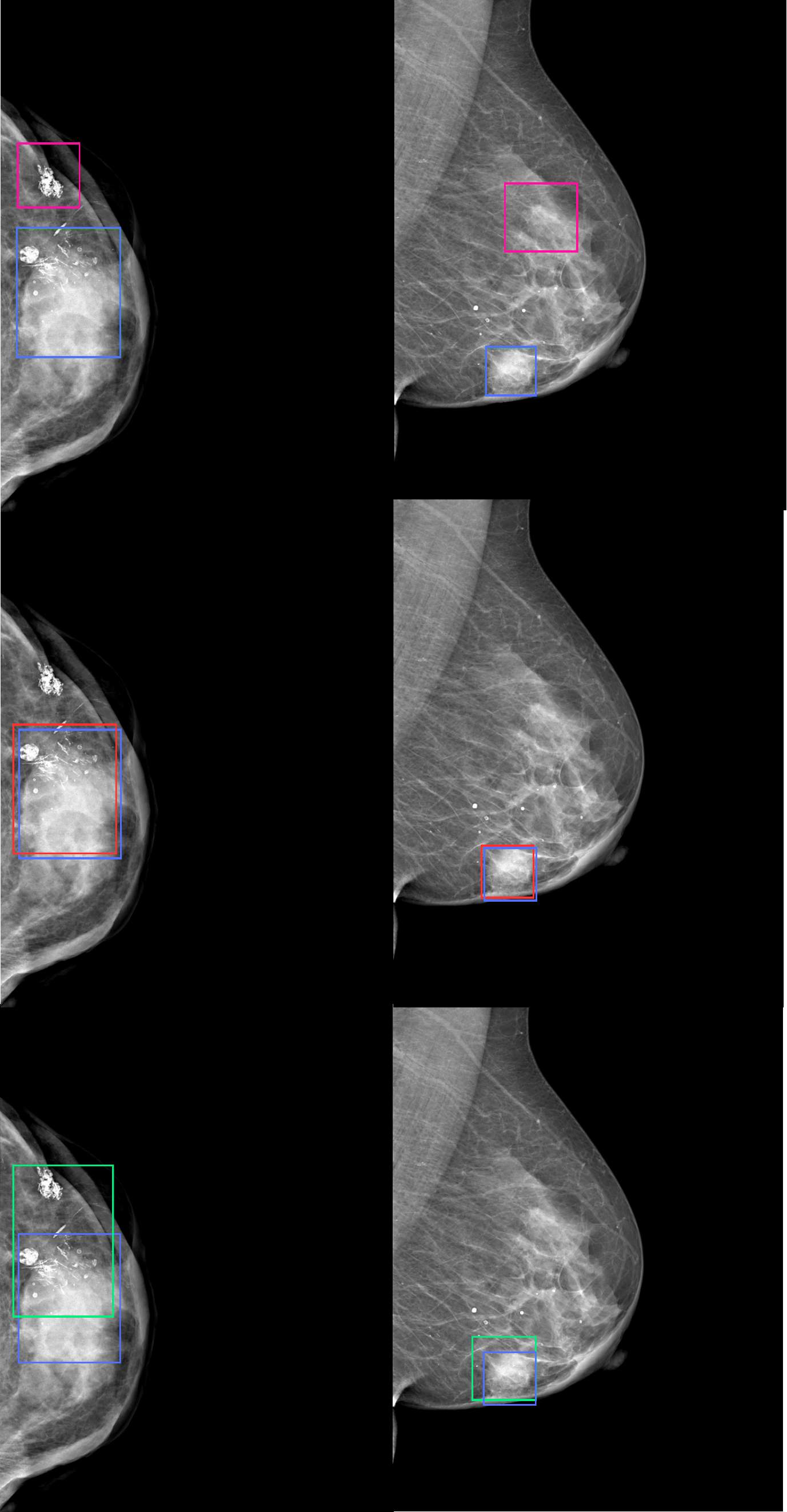}
        \caption{RSNA $\rightarrow$ INB}
        \label{fig:rsna2inb}
    \end{subfigure}
    \caption{Qualitative results of our proposed Grounded Teacher on Medical settings (DDSM $\rightarrow$ RSNA and RSNA $\rightarrow$ INBreast). We present visual comparisons respectively between MRT~\cite{zhao2023masked} (top Row), AT~\cite{li2022cross} , GT (bottom row). Our method, GT, effectively identifies Malignancies better and mitigates false negatives. Bounding box colors indicate:  Blue — Ground truth, Green — true positive, Pink -- False Positive, and Red — false negative.}
    \label{fig:main}
\end{figure}

\begin{figure}[t]
    \centering
    \includegraphics[width=\columnwidth]{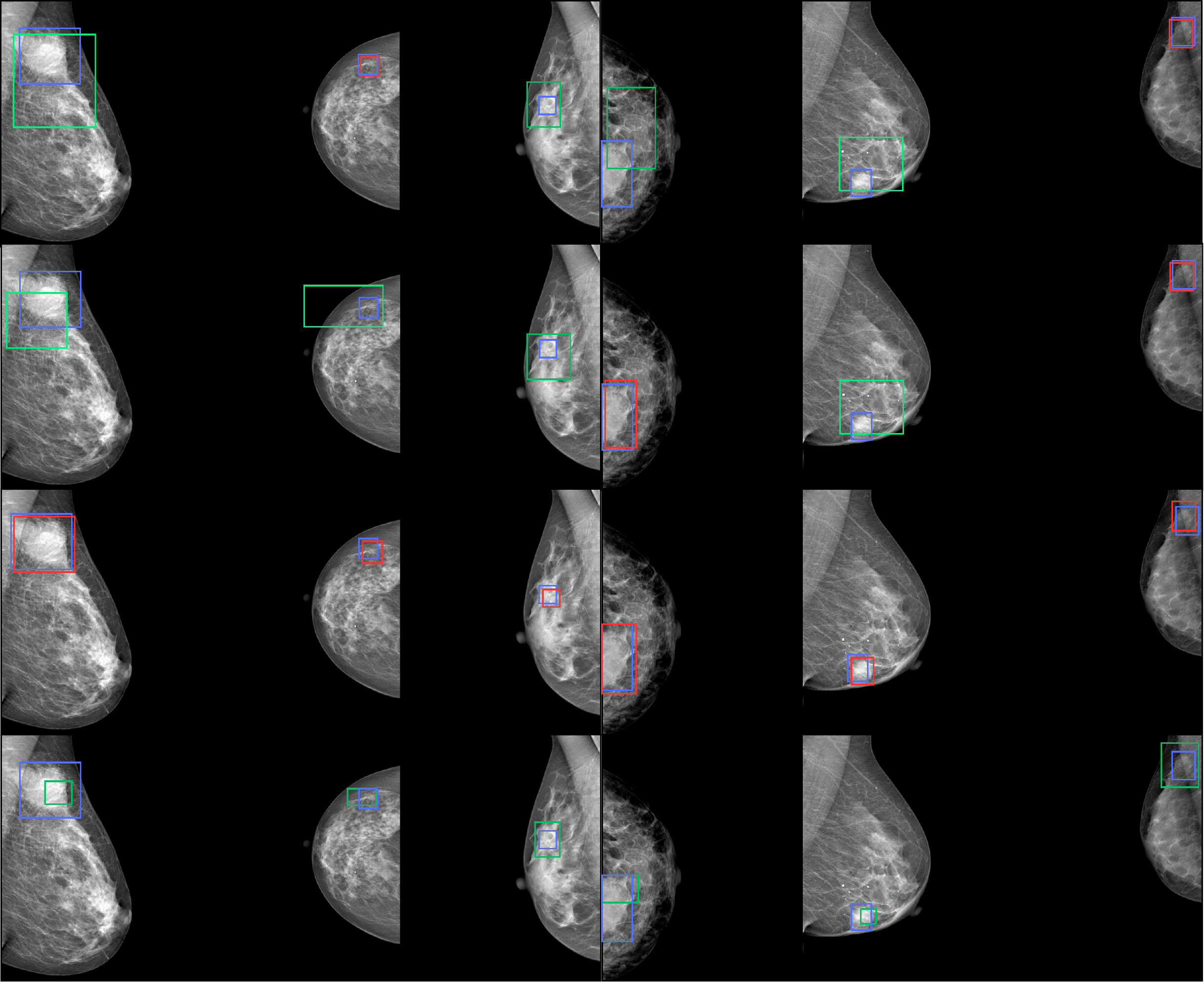}
    \caption{Qualitative results of our proposed Grounded Teacher on Medical settings (DDSM $\rightarrow$ INBreast). We present visual comparisons respectively between MRT~\cite{zhao2023masked} (top Row), D-Master~\cite{ashraf2024dmastermaskannealedtransformer}, AT~\cite{li2022cross}, GT (bottom row). Our method, GT, effectively identifies Malignancies better and mitigates false negatives. Bounding box colors indicate:  Blue — Ground truth, Green — true positive, Pink -- False Positive, and Red — false negative.}
    \label{fig:ddsm2inb}
\end{figure}

%

\subsubsection{Qualitative Results}

Figures~\ref{fig:rsna2inb},~\ref{fig:ddsm2inb},~\ref{fig:ddsm2rsna}, and Figure~\ref{qual_all} present the updated qualitative results of our proposed method on both medical and natural domain shifts, respectively. In natural settings (Cityscapes $\rightarrow$ Foggy Cityscapes; Figure~\ref{qual_all}), we compare predictions from competing approaches (MRT, IRG, LPLD, and AT; top rows) with those from the proposed Grounded Teacher (\gt) shown in the bottom row. The \gt effectively corrects misclassifications (blue boxes) and substantially reduces false positives and false negatives (pink and red boxes), particularly under adverse weather-induced domain shift. Notably, improvements are evident across object scales, including small and partially occluded objects, demonstrating enhanced localization robustness.
In medical settings (Figure~\ref{fig:rsna2inb},~\ref{fig:ddsm2inb},~\ref{fig:ddsm2rsna}), we observe consistent qualitative improvements across all evaluated domain shifts (DDSM $\rightarrow$ RSNA, RSNA $\rightarrow$ INBreast, and DDSM $\rightarrow$ INBreast). Compared to MRT, AT, and D-MASTER, the \gt produces tighter and more anatomically consistent bounding boxes while recovering missed lesions, especially in low-contrast and small-lesion cases. These visual improvements directly align with the newly added localization-aware evaluation reported in Table~\ref{tab:medical_fpi_all} (Appendix), where \gt consistently reduces false negatives at fixed false positives per image (FPI), particularly in low-FPI regimes.
Overall, the updated qualitative results in both natural and medical domains, together with the new localization tables and FROC analysis (Figure~\ref{fig:froc_plots}), confirm that the performance gains of \gt stem from improved localization-aware generalization under domain shift rather than recall-only behavior.

\section{Limitations}

We also evaluated the Cityscapes → BDD100k setup to test further on diverse domains. However, we observed subpar performance on this setup due to several inherent challenges. First, BDD100k is significantly larger with 100k annotated images labeled object instances across 10 categories \cite{bdd100k}. In contrast, Cityscapes contains fewer than 3k training images with fewer than 8 classes and more consistent urban scenes. The increased scale, greater intra-class diversity, and higher number of object categories in BDD100k likely introduce a domain gap too wide for effective unsupervised adaptation under our current framework. Our architecture, when trained on Cityscapes and tested directly on BDD100k, does not surpass existing state-of-the-art (\sota) \sfod methods, but it does show improvement with adaptation over source-only baselines as noted in Table~\ref{table:c2b_filtered}. 

\begin{table}[t]
\setlength\tabcolsep{3pt}
\renewcommand{\arraystretch}{1.1}
\centering
\begin{tabularx}{\linewidth}{@{}l|>{\centering\arraybackslash}X@{}}
\toprule
\textbf{Method} & \textbf{mAP} \\
\midrule
\texttt{Source-trained} & 20.2 \\
\texttt{SFOD}~\cite{li2020free} & 26.6 \\
\texttt{SFOD-Mosaic}~\cite{li2020free} & 29.0 \\
\texttt{HCL}~\cite{huang2022model} & \textbf{30.3} \\
\rowcolor[gray]{0.9}
\gt (Ours) & \underline{29.1} \\
\bottomrule
\end{tabularx}
\caption{Source-free adaptation from Cityscapes→BDD100k (\texttt{CB}). GT generalizes beyond medical imagery to structured natural scenes and performs competitively with recent SFOD baselines.}
\label{table:c2b_filtered}
\end{table}

Considering our motivation and primary contribution lie in adapting models for medical image analysis, our method achieves promising results across multiple breast cancer detection datasets, validating its effectiveness in the domain. The challenges observed in the Cityscapes → BDD100k setup while the simultaneous improvement with adaptation over source-only baselines, highlight opportunities for future work regarding which, based on our findings, we believe that further investigation into inter-class dynamics is a promising direction for advancing context imbalance in the \sfod setting as a whole, especially in medical scenarios.

\section{Conclusion}

We have proposed a novel source-free domain adaptive object detection framework that leverages a Relation Contextual module and Expert Branch, effectively approximating and mitigating class biases, leading to more equitable performance across classes.
Through comprehensive experiments conducted across various natural and medical image benchmark datasets, we consistently demonstrate that our technique outperforms the state-of-the-art. Our ablation studies confirm the effectiveness of each design aspect in improving the model's domain adaptation capability.

\noindent\textbf{Data Availability Statement.} 
All results presented in this paper are based on publicly available datasets. To ensure reproducibility, we are making the processed data openly available. The datasets, along with the associated processing scripts, will be publicly released at the following GitHub repository at \href{https://github.com/Tajamul21/Grounded_Teacher}{this link}. This repository will assist other researchers in replicating the results and conducting further experiments.

\clearpage

\bibliographystyle{plain}
\bibliography{sn-bibliography-2}

\clearpage
\onecolumn
\section*{Appendix}\label{secA1}

\subsection*{Further Details on Datasets.}
Our experiments are conducted on six datasets. We also introduce the \sfod problem in the medical domain and conduct experiments on three Breast Cancer Detection (\bcd) datasets.

1) \textbf{Cityscapes}~\cite{cordts2016cityscapes} is gathered from urban environments across 50 European cities, provides detailed annotations for 30 semantic classes across 8 categories. It comprises 5,000 high-quality annotated images and a larger set of 20,000 coarsely annotated images, all high-resolution (2048x1024 pixels) providing detailed visual data for precise scene understanding. Our study utilizes the high-quality subset of 5,000 images, consisting of 2,975 training images and a standard test set of 500 images from Frankfurt, Munster, and Lindau, as employed in prior research. This dataset is release under the Creative Commons Attribution-NonCommercial-ShareAlike 4.0 International License, and is available for academic and research purposes.

2) \textbf{Foggy Cityscapes}~\cite{sakaridis2018semantic} is an extension of the Cityscapes dataset, designed to support research in developing robust computer vision algorithms for autonomous driving in foggy conditions. Similarly, it consists of high-resolution images (2048x1024 pixels) with annotations ofr 2D bounding boxes, pixel-level semantic segmentation, and instance segmentation, inherited from the original dataset. It is directly constructed from Cityscapes by simulating three levels of foggy weather (0.005,0.001,0.02), but we adapt on the most extreme level (0.02) in our experiments as done by Kennerley et al. \cite{kennerley2024catexploitinginterclassdynamics}. This dataset is released under the Creative Commons Attribution-NonCommercial-ShareAlike 4.0 International License.

3) \textbf{INBreast}~\cite{moreira2012inbreast} is a relatively small breast cancer detection dataset, consisting of 410 mammography images from 115 patients, including 87 confirmed malignancies. The training images include both histologically confirmed cancers and benign lesions initially recalled for further examination but later identified as nonmalignant. We hypothesize that incorporating both malignant and benign lesions in the training process will enhance our model's ability to detect a broader range of lesions and effectively distinguish between malignant and benign cases.

\begin{table*}[thbp]
\caption{Details of Training Datasets}
\label{tab:training_datasets}
\centering
\resizebox{\textwidth}{!}{%
\begin{tabular}{c|c|c|c|p{2.2 cm}|c}
\toprule
\textbf{Dataset} & \textbf{Type} & \multicolumn{2}{c|}{\textbf{Pre-training (Source)}} & \multicolumn{2}{c}{\textbf{Unsupervised Adaptation (Target)}} \\
\midrule
& & \textbf{Train} & \textbf{Val/Test} & \centering\textbf{Train}  & \textbf{Test} \\ \midrule
Cityscapes & Natural & 2,975 & 500 & \centering 2,975  & 500 \\ 
Foggy Cityscapes & Natural & - & - & \centering 2,975 & 500 \\ 
BDD100k & Natural &  - & - & \centering 36,728 & 5,258 \\ \midrule
RSNA & Medical & 895 (170) & 98 (18) &\centering 993 (188) & 993(188) \\ 
INBreast & Medical & - & - &\centering 410 (91) & 410 (91) \\ 
DDSM & Medical & 2885 (1339) & 218 (118) & \centering3103 (1458) &3103 (1458) \\ 

\bottomrule
\end{tabular}
}
\end{table*}

4) \label{section:DDSM}\textbf{DDSM}~\cite{lee2017curated} is a publicly available breast cancer detection dataset, comprising 2,620 full mammography images, with 1,162 containing malignancies. The DDSM dataset offers digitized film-screen mammography exams with lesion annotations at the pixel level, where cancerous lesions are histologically confirmed. We used the DDSM dataset exclusively for training our model and not for evaluation. This decision stems from the observation that the quality of digitized film-screen mammograms is inferior to that of full-field digital mammograms, making evaluation on these cases less relevant. For our purposes, we converted the lossless \jpeg images to \png format, mapped pixel values to optical density using calibration functions from the DDSM website, and rescaled the pixel values to a range of 0–255.

5) \textbf{RSNA-BSD1K}  The RSNA dataset \cite{carr2022rsna} was contributed by mammography screening programs in Australia and the U.S. It contains radiographic breast images of female subjects in dicom format. It includes detailed labels, with radiologists’ evaluations and follow-up pathology results for suspected malignancies. The original RSNA dataset consists of 54,706 screening mammograms, containing 1,000 malignancies from 8,000 patients.  We curated a subset named RSNA-BSD1K, comprising 1,000 mammograms with 200 malignant cases, annotated at the bounding box level by 2 expert radiologists.

5) \textbf{BDD100k} The BDD100k dataset \cite{bdd100k}, created by researchers primarily at UC Berkeley to address the limitations of prior driving datasets in scale and diversity, provides 100,000 diverse, 720p driving videos (split 70K train/10K val/20K test) from varied US locations, weather, and times. For object detection, one frame per video (at the 10th second) is annotated with bounding boxes for 10 categories, totaling 100K annotated images. It features over 1.8 million annotated instances across categories like car (1M+), sign (343K+), light (265K+), and person (129K+). Visibility attributes indicate about 48\% of instances are occluded and 7\% truncated. Its scale (70K training images) and diversity support robust detector training and domain adaptation studies, and significantly boost performance in related tasks like instance segmentation and tracking when used in joint training.

It is important to note that unlike single domain detection techniques for natural images, where a subset of the target dataset is reserved for adaptation and another for testing, our approach operates without requiring any labeled samples from the target domain. This characteristic makes it particularly suitable for medical datasets, where it is more practical to utilize the entire dataset during both training and testing. Therefore, when we refer to training on “Dataset A to Dataset B,” we imply that the model is trained using the full source dataset ($\mathcal{D}_s = A$) and adapted to the full target dataset ($\mathcal{D}_t = B$) in a fully unsupervised manner. Table~\ref{tab:training_datasets} outlines the specific dataset splits used in our experiments.

\begin{table*}[t]
\centering
\caption{Data Augmentation Methods and Parameters}
\label{tab:augmentation}
\begin{tabular}{c|c|c}
\toprule
\textbf{Augmentation Type} & \textbf{Method} & \textbf{Parameters} \\ \midrule
\multirow{3}{*}{Weak} 
  & Random Horizontal Flip & Probability = 0.5 \\ \cmidrule{2-3}
  & Resize & Size = 800, Max Size = 1333 \\ \midrule
\multirow{8}{*}{Strong} 
  & \multirow{4}{*}{Color Jitter (Color Adjustment)} 
  & Brightness = 0.4 \\ \cmidrule{3-3}
  & & Contrast = 0.4 \\ \cmidrule{3-3}
  & & Saturation = 0.4 \\ \cmidrule{3-3}
  & & Hue = 0.1 \\ \cmidrule{2-3}
  & Random Grayscale & Probability = 0.2 \\ \cmidrule{2-3}
  & \multirow{2}{*}{Gaussian Blur} 
  & Sigma Range = [0.1, 2.0 \\ \cmidrule{3-3}
  & & Probability = 0.5\\ \cmidrule{2-3}
  & \multirow{2}{*}{Normalization} 
  & Mean = [0.485, 0.456, 0.406] \\ \cmidrule{3-3}
  & & Std = [0.229, 0.224, 0.225] \\ \bottomrule
\end{tabular}%
\end{table*}

\subsection*{More Details on Augmentations}
\label{S_augmentation}

In our study, we employed both weak and strong augmentation techniques to enhance the robustness and generalization capabilities of our model. These augmentations are applied to the training images to simulate various real-world scenarios and improve the model's performance. The detailed data augmentation and parameters are shown in Table~\ref{tab:augmentation}.

\noindent\textbf{Weak Augmentation. }
Weak augmentations are relatively simple transformations that slightly alter the images without significantly changing their content. We utilized two primary weak augmentation techniques. The first is \textit{Random Horizontal Flip}, which flips the image horizontally with a probability of 0.5. This helps the model learn invariance to the left-right orientation of objects, making it more robust to such variations. The second technique is \textit{Resize}, where images are resized such that the shortest edge is 800 pixels while maintaining the aspect ratio, and if the longest edge exceeds 1333 pixels, the image is scaled down accordingly. This resizing standardizes the input image size, ensuring consistent and efficient training. Together, these weak augmentations provide a baseline level of variability in the training data, helping the model to generalize better across different image scales and orientations.

\noindent\textbf{Strong Augmentation. }
Strong augmentations involve more complex transformations that significantly alter the images, thereby providing a broader range of variability. These augmentations are designed to challenge the model and improve its ability to handle diverse and complex real-world scenarios. The first strong augmentation is \textit{Color Jitter}, which randomly changes the brightness, contrast, saturation, and hue of the images with specified parameters (brightness = 0.4, contrast = 0.4, saturation = 0.4, hue = 0.1) and a probability of 0.8. This helps the model become invariant to different lighting conditions and color variations. The second augmentation is \textit{Random Grayscale}, which converts the image to grayscale with a probability of 0.2, encouraging the model to focus on shapes and structures rather than colors. The third strong augmentation is \textit{Gaussian Blur}, applied with a sigma range of [0.1, 2.0] and a probability of 0.5, simulating out-of-focus conditions and reducing high-frequency noise. Finally, all images are converted to tensors and normalized using the mean [0.485, 0.456, 0.406] and standard deviation [0.229, 0.224, 0.225]. These strong augmentations introduce substantial variability in the training data, forcing the model to learn more robust and generalized features.

\subsection*{Details on Hyperparameters}

\begin{table*}[t]
    \centering
    \caption{Detailed hyper-parameters corresponding to each benchmark for Medical settings (DDSM $\rightarrow$ INBreast) and Natural settings (Cityscapes $\rightarrow$ Foggy Cityscapes)}
    \vspace{0.1 em}
    \label{tab:hyperparams}
    \begin{adjustbox}{width=0.95\textwidth}
    \begin{tabular}{@{} cccc @{}}
        \toprule
        \textbf{Hyper-parameter} & \textbf{Description} & \textbf{DDSM $\rightarrow$ INBreast} & \textbf{Cityscapes $\rightarrow$ Foggy Cityscapes} \\
        \midrule
        - & Detector & FRCNN & FRCNN \\
        - & Backbone & VGG  & VGG  \\
        - & BatchNorm & True & True \\
        - & Expert & BioMedParse \cite{Zhao_2024} & GroundingDINO \cite{liu2024groundingdinomarryingdino} \\
        $N_c$ & Number of categories for classification head & 1 & 8 \\
        $\alpha$ & Decay rate for student-teacher EMA & 0.9996 & 0.9996 \\
        $\beta$ & Beta-distribution parameters for mixup & [0.5,0.5] & [0.5,0.5] \\
        $\lambda_d$ & Weight for Adversarial Loss & 0.1 & 0.1 \\
        $\lambda_u$ & Weight for Student Loss & 1.0 & 1.0 \\
        $\lambda_e$ & Weight for Expert Loss & 1.0 & 1.0 \\
        $\tau$ & Threshold value for pseudo-label confidence & 0.8 & 0.8 \\
        $\lambda_I$ & Regularization term for Semantic Aware Loss \texttt{SAL} & 1.0 & 1.0 \\
        - & Source-similar Augmentation Ratio & 0.5 & 0.5 \\
        - & Source-dissimilar Augmentation Ratio & 0.5 & 0.5 \\
        - & Burn-Up Step Iterations & 20000 & 20000 \\
        - & Total Training Iterations & 40000 & 80000 \\
        $base\_lr$ & Learning Rate & 0.01 & 0.02 \\
        \bottomrule
    \end{tabular}
    \end{adjustbox}
\end{table*}

The hyperparameters used for training the models on the Medical setting (DDSM $\rightarrow$ INBreast) and Natural setting (Cityscapes $\rightarrow$ Foggy Cityscapes) domain adaptation task are summarized in Table~\ref{tab:hyperparams}. We employ the Adam optimizer with a base learning rate of $1 \times 10^{-4}$ and a weight decay of $1 \times 10^{-5}$ to ensure stable training. The models are trained for 50 epochs with a batch size of 16, and the input image resolution is fixed to $512 \times 512$ pixels. 

To balance the contributions from different training objectives, we set the unsupervised loss weight $\lambda_u$ to 2.0 and the discriminator loss weight $\lambda_d$ to 0.5. The expert loss is composed of a classification and regression component, both weighted equally with $\lambda_{\text{cls}} = 1.0$ and $\lambda_{\text{reg}} = 1.0$. For data augmentation, we apply standard techniques including random flipping, scaling, and color jittering to improve generalization. 

These hyperparameters were found to be effective for domain adaptation under adverse weather conditions, and we retain them consistently across all comparative experiments unless explicitly varied in ablation studies.

\subsection*{Localization Analysis under Domain Shift}

Accurate localization is critical for reliable object detection, particularly in medical imaging scenarios where imprecise bounding boxes or missed detections may lead to incorrect clinical interpretation. In response to the reviewer’s concern regarding localization quality and recall-dominant behavior, we provide a detailed localization-focused analysis using the newly added false positives per image (FPI) evaluation in Table~\ref{tab:medical_fpi_all}.
Table~\ref{tab:medical_fpi_all} reports detection statistics at fixed FPIs across three challenging medical domain shifts. At low-FPI operating points (0.05 and 0.30), which are especially relevant for clinical deployment, our source-free method consistently reduces false negatives while also controlling false positives more effectively than competing approaches. For example, in the DDSM $\rightarrow$ INBreast setting at 0.05 FPI, our method reduces false negatives from 9 (source and MRT) to 8, while simultaneously lowering false positives compared to the source baseline (FP: 2 vs.\ 3). At 0.30 FPI, our approach achieves fewer false negatives (5) than both the source (7) and MRT (6), while also operating at a lower false positive count.

Importantly, these improvements are not achieved by inflating false positives. As shown in Table~\ref{tab:medical_fpi_all}, our method often operates at lower FP counts than both the source and MRT models, especially at clinically relevant low-FPI points. At higher FPIs (0.50 and 1.00), our approach continues to preserve competitive false positive rates while consistently reducing missed detections, further confirming robust localization under severe domain shift.
The quantitative trends in Table~\ref{tab:medical_fpi_all} are strongly supported by the qualitative results presented in Figures~\ref{fig:rsna2inb}, \ref{fig:ddsm2inb}, \ref{fig:ddsm2rsna}, and Figure~\ref{qual_all}, where our method produces tighter and more anatomically consistent bounding boxes while eliminating spurious detections. In particular, small and low-contrast lesions that are frequently missed or poorly localized by competing approaches are correctly detected and localized by our method. Together, these results demonstrate that the observed performance gains stem from improved localization-aware generalization under domain shift, rather than recall-only behavior.

\begin{table}[b]
\centering
\caption{Detection performance across medical domain shifts at different false positives per image (FPI).}
\label{tab:medical_fpi_all}

\renewcommand{\arraystretch}{1.3}
\setlength{\tabcolsep}{8pt}
\rowcolors{2}{blue!5}{white}

\begin{tabular}{l lccc|ccc|ccc}
\toprule
\rowcolor{blue!20}
Task & FPI & \multicolumn{3}{c|}{Ours (Source Free)} & \multicolumn{3}{c|}{Source Method} & \multicolumn{3}{c}{MRT (with source)} \\
\rowcolor{blue!10}
 &  & TP & FP & FN & TP & FP & FN & TP & FP & FN \\
\midrule

\multicolumn{11}{l}{\textbf{DDSM $\rightarrow$ INBreast}} \\
\midrule
 & 0.05 & 1 & 2 & 8 & 0 & 3 & 9 & 0 & 2 & 9 \\
 & 0.30 & 4 & 10 & 5 & 2 & 13 & 7 & 3 & 13 & 6 \\
 & 0.50 & 6 & 17 & 3 & 5 & 24 & 4 & 4 & 21 & 5 \\
 & 1.00 & 8 & 36 & 1 & 8 & 40 & 1 & 6 & 38 & 3 \\

\midrule
\multicolumn{11}{l}{\textbf{RSNA $\rightarrow$ INBreast}} \\
\midrule
 & 0.05 & 0 & 1 & 8 & 0 & 3 & 11 & 0 & 2 & 9 \\
 & 0.30 & 3 & 10 & 6 & 3 & 15 & 6 & 2 & 13 & 7 \\
 & 0.50 & 4 & 18 & 5 & 5 & 25 & 4 & 3 & 22 & 6 \\
 & 1.00 & 8 & 37 & 1 & 8 & 40 & 1 & 6 & 39 & 3 \\

\midrule
\multicolumn{11}{l}{\textbf{DDSM $\rightarrow$ RSNA}} \\
\midrule
 & 0.05 & 2 & 4 & 17 & 0 & 5 & 19 & 1 & 5 & 18 \\
 & 0.30 & 8 & 27 & 11 & 6 & 32 & 13 & 6 & 29 & 13 \\
 & 0.50 & 11 & 49 & 8 & 10 & 50 & 9 & 9 & 52 & 10 \\
 & 1.00 & 18 & 105 & 1 & 18 & 107 & 1 & 14 & 105 & 5 \\

\bottomrule
\end{tabular}
\end{table}

\newpage
\subsection*{Additional Visualizations}

\begin{figure}[htbp] 
    \centering 

    \begin{subfigure}{\textwidth} 
        \centering 
        \includegraphics[width=\textwidth]{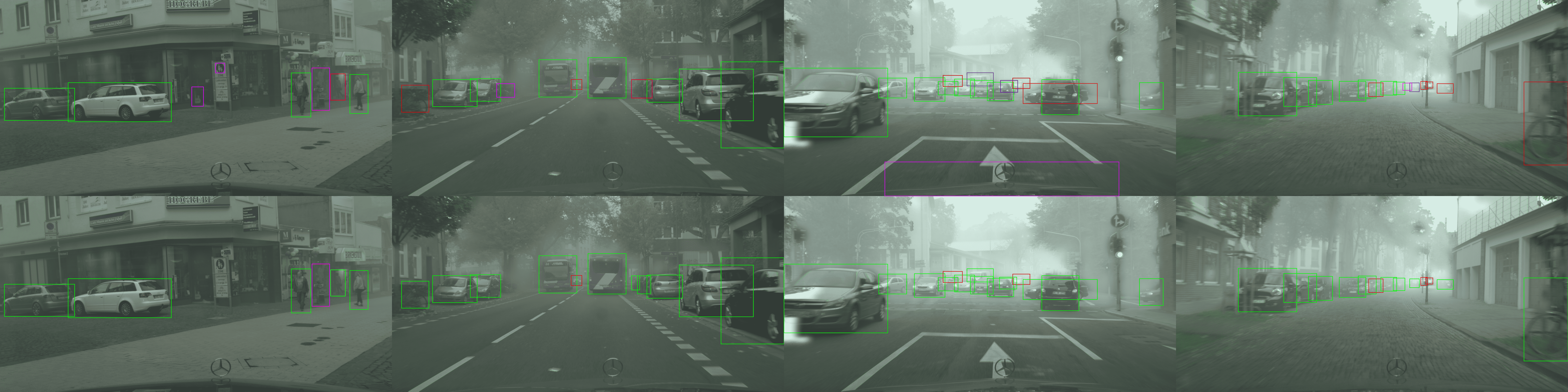}
        \subcaption{\textbf{Aachen}} 
        \label{fig:qual_aachen} 
    \end{subfigure}

    \begin{subfigure}{\textwidth}
        \centering
        \includegraphics[width=\textwidth]{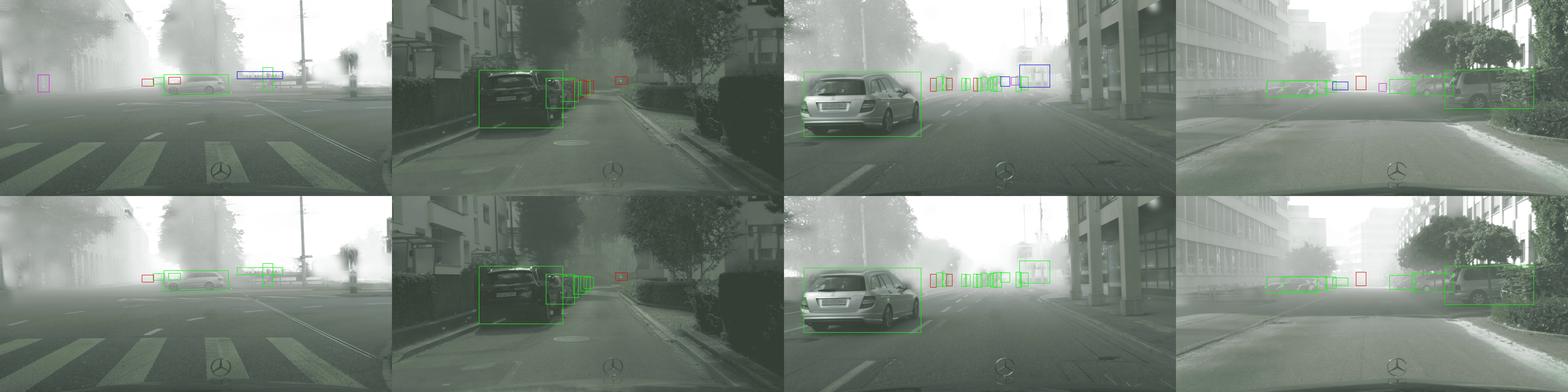}
        \subcaption{\textbf{Zurich}}
        \label{fig:qual_zurich}
    \end{subfigure}

    \begin{subfigure}{\textwidth}
        \centering
        \includegraphics[width=\textwidth]{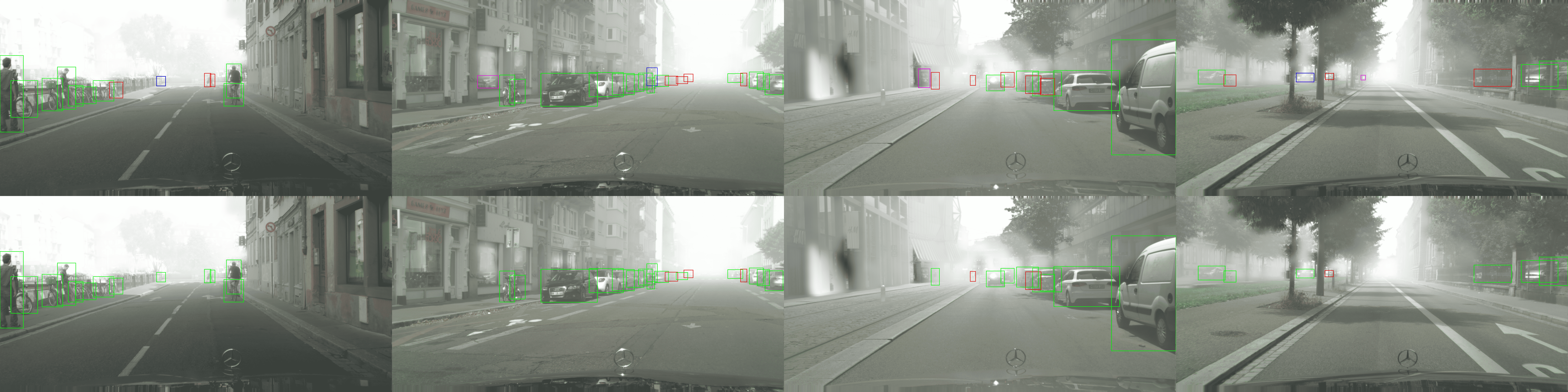}
        \subcaption{\textbf{Strasbourg}}
        \label{fig:qual_strasbourg}
    \end{subfigure}

    \caption{Qualitative results comparing our proposed Grounded Teacher (\gt) with the baseline (\at) on Natural settings (Cityscapes $\rightarrow$ Foggy Cityscapes) using challenging foggy driving scenes from (a) Aachen, (b) Zurich, and (c) Strasbourg. Each sub-figure presents the \at output (top row) and the corresponding \gt output (bottom row). Our method (\gt) demonstrates improved performance by mitigating misclassifications, reducing false negatives, and eliminating false positives compared to \at. Bounding box colors indicate detection status: Green — true positive, Blue — misclassified, Red — false negative, Pink — false positive.}
    \label{fig:qual_all} 
\end{figure}

\begin{figure}[htbp] 
    \centering 

    \begin{subfigure}{\textwidth} 
        \centering 
        \includegraphics[width=\textwidth]{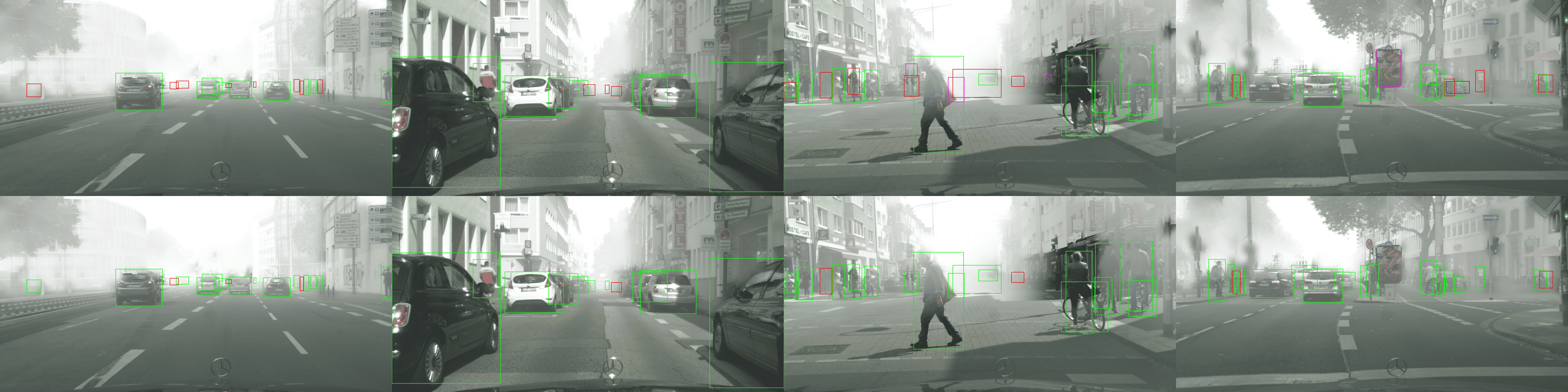}
        \subcaption{\textbf{Cologne}} 
        \label{fig:qual_aachen} 
    \end{subfigure}

    \begin{subfigure}{\textwidth}
        \centering
        \includegraphics[width=\textwidth]{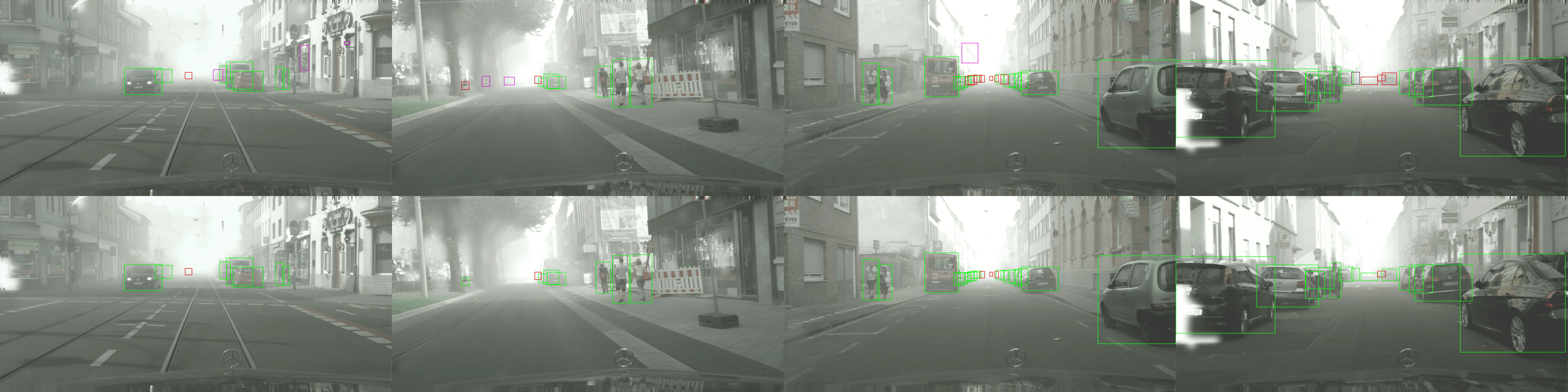}
        \subcaption{\textbf{Krefeld}}
        \label{fig:qual_zurich}
    \end{subfigure}

    \begin{subfigure}{\textwidth}
        \centering
        \includegraphics[width=\textwidth]{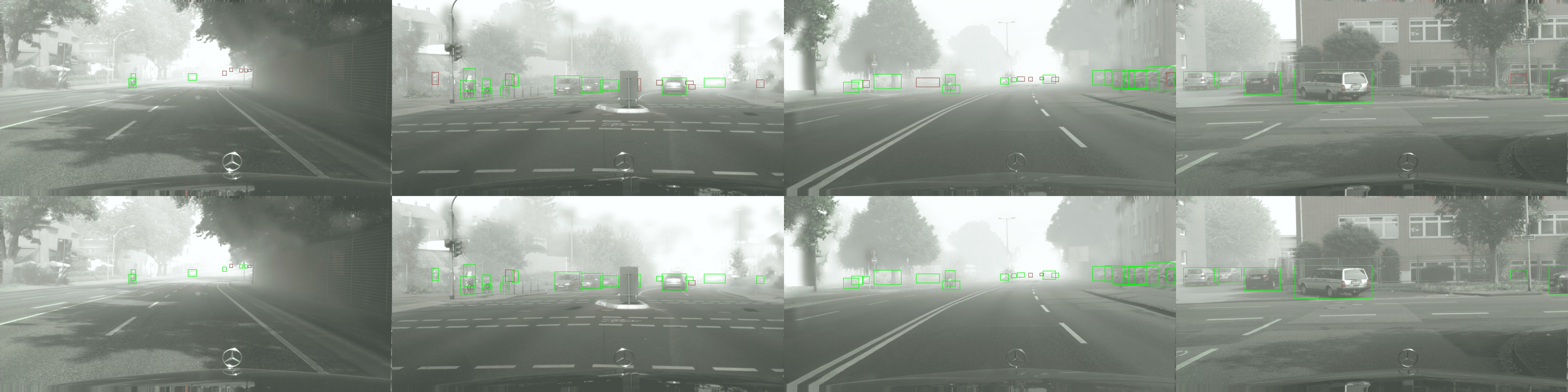}
        \subcaption{\textbf{Monchengladbach}}
        \label{fig:qual_strasbourg}
    \end{subfigure}

    \caption{(Contd.) Qualitative results comparing our proposed Grounded Teacher (\gt) with the baseline (\at) on Natural settings (Cityscapes $\rightarrow$ Foggy Cityscapes) using challenging foggy driving scenes from (a) Cologne, (b) Krefeld, and (c) Monchengladbach. Each sub-figure presents the \at output (top row) and the corresponding \gt output (bottom row). Our method (\gt) demonstrates improved performance by mitigating misclassifications, reducing false negatives, and eliminating false positives compared to \at. Bounding box colors indicate detection status: Green — true positive, Blue — misclassified, Red — false negative, Pink — false positive.}
    \label{fig:qual_all} 
\end{figure}

\begin{figure}[htbp] 
    \centering 

    \begin{subfigure}{\textwidth} 
        \centering 
        \includegraphics[width=\textwidth]{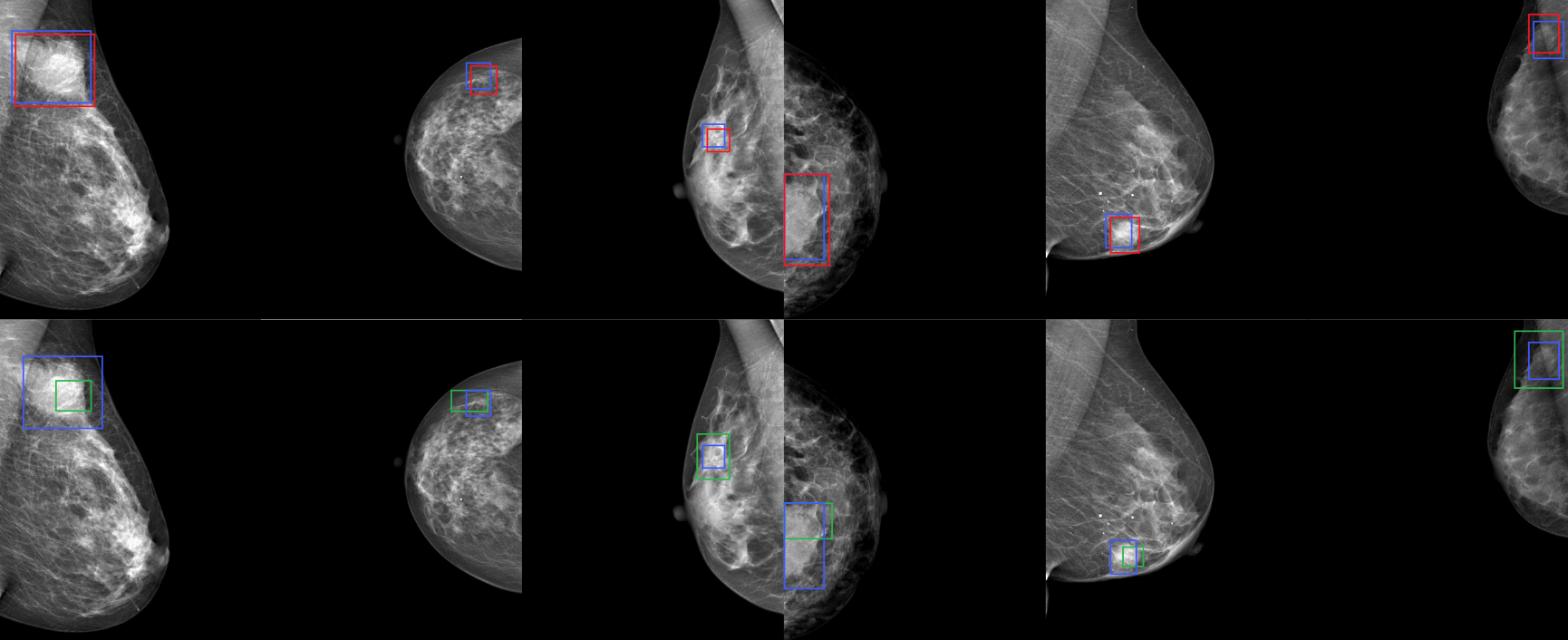}
        \subcaption{\textbf{DDSM $\rightarrow$ INBreast}} 
        \label{fig:qual_aachen} 
    \end{subfigure}

    \begin{subfigure}{\textwidth}
        \centering
        \includegraphics[width=\textwidth, height=165px]{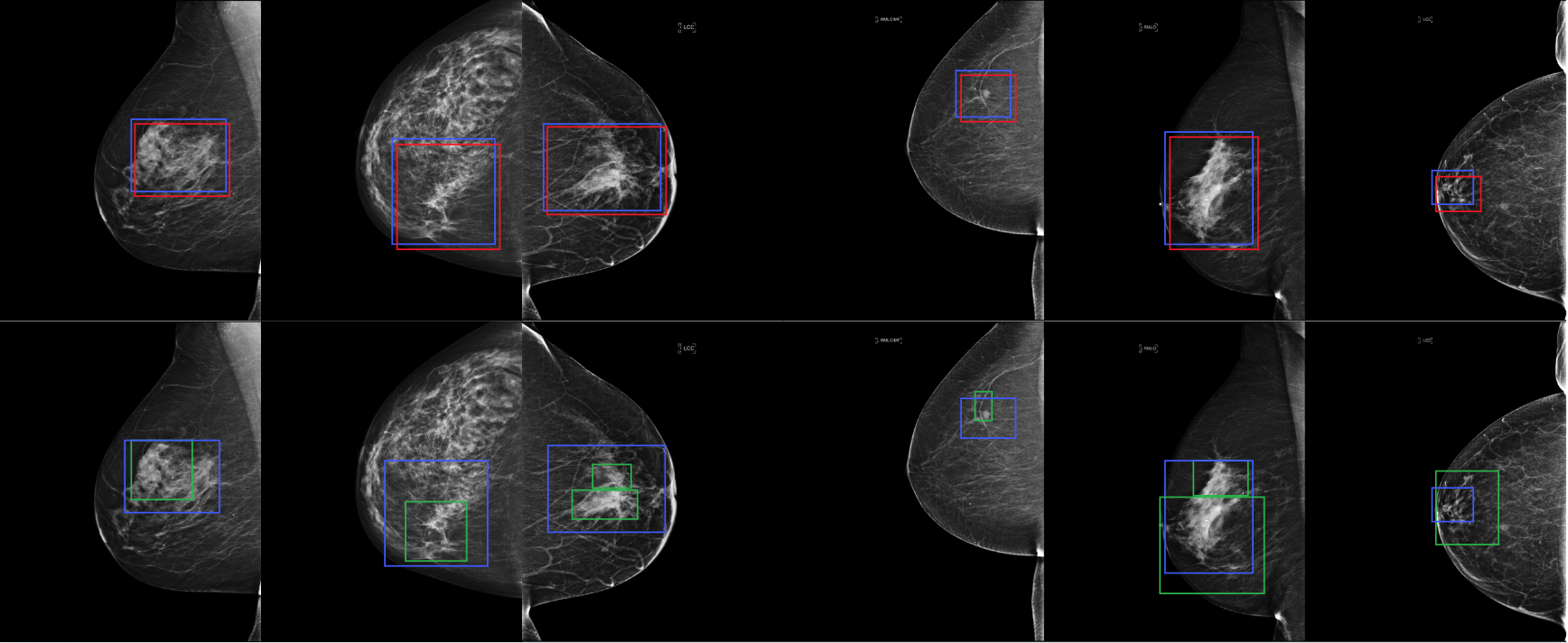}
        \subcaption{\textbf{DDSM $\rightarrow$ RSNA}}
        \label{fig:qual_zurich}
    \end{subfigure}

    \begin{subfigure}{\textwidth}
        \centering
        \includegraphics[width=\textwidth, height=165px]{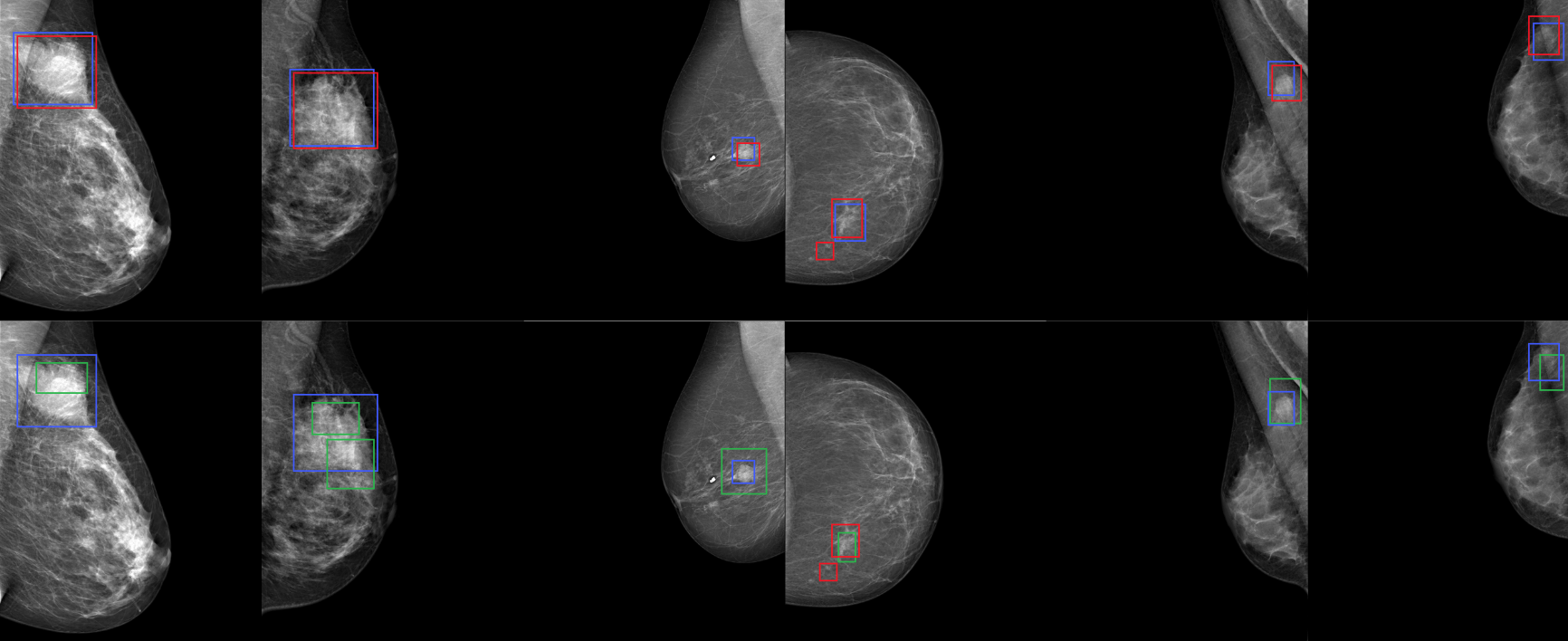}
        \subcaption{\textbf{RSNA $\rightarrow$ INBreast}}
        \label{fig:qual_strasbourg}
    \end{subfigure}

    \caption{Qualitative results comparing our proposed Grounded Teacher (\gt) with the baseline (\at) on Medical settings using challenging Mammograms adapted from (a) DDSM $\rightarrow$ INBreast, (b) DDSM $\rightarrow$ RSNA, and (c) RSNA $\rightarrow$ INBreast. Each sub-figure presents the \at output (top row) and the corresponding \gt output (bottom row). Our method (\gt) demonstrates improved performance by effective identification of Malignancies and mitigating false negatives compared to \at. Bounding box colors indicate detection status: Blue — Ground truth, Green — true positive, Red — false negative.}
    \label{fig:qual_all} 
\end{figure}

\end{document}